\documentclass{article}

\usepackage{microtype}
\usepackage{graphicx}
\usepackage{subfigure}
\usepackage{amsfonts}
\usepackage{amsmath}
\usepackage{booktabs} 

\usepackage{algorithm}
\usepackage{algorithmic}

\usepackage{hyperref}

\usepackage{titlesec}

\usepackage[accepted]{icml2024}

\icmltitlerunning{Dynamic Anisotropic Smoothing}

\begin{document}

\twocolumn[
\icmltitle{Dynamic Anisotropic Smoothing for Noisy Derivative-Free Optimization}




\begin{icmlauthorlist}
\icmlauthor{Sam Reifenstein$^{*}$}{NTTRI}
\icmlauthor{Timothee Leleu$^{*}$}{NTTRI,Stanford}
\icmlauthor{Yoshihisa Yamamoto}{NTTRI}
\end{icmlauthorlist}

\icmlaffiliation{NTTRI}{NTT Research Inc}
\icmlaffiliation{Stanford}{Stanford University}
\icmlcorrespondingauthor{Sam Reifenstein}{Sam.Reifenstein@ntt-research.com}

\icmlkeywords{Machine Learning, ICML}

\vskip 0.3in
]



\printAffiliationsAndNotice{\icmlEqualContribution} 

\begin{abstract}

We propose a novel algorithm that extends the methods of ball smoothing and Gaussian smoothing for noisy derivative-free optimization by accounting for the heterogeneous curvature of the objective function. The algorithm dynamically adapts the shape of the smoothing kernel to approximate the Hessian of the objective function around a local optimum. This approach significantly reduces the error in estimating the gradient from noisy evaluations through sampling. We demonstrate the efficacy of our method through numerical experiments on artificial problems. Additionally, we show improved performance when tuning NP-hard combinatorial optimization solvers compared to existing state-of-the-art heuristic derivative-free and Bayesian optimization methods.

\end{abstract}

\section{Introduction}

\subsection{Problem formulation}

The problem of optimizing an objective function without access to its gradient, also known as derivative-free optimization, is a well-developed field due to its widespread applications\cite{Larson2019,Berahas2022,Gasnikov2022R}. We consider derivative-free optimization problems of the form:
\begin{align}
\text{argmax}_{x \in \mathbb{R}^D} f(x)
\end{align}



where an oracle gives us access to only noisy evaluation $\hat{f}$ to the objective function $f$ with $\hat{f}(x, \zeta) = f(x) + \zeta(x)$ and $\zeta$ some noise term which can be deterministic or stochastic bounded noise such that $\zeta(x) < \zeta_f$. In particular, the gradient $\nabla_x f$ of $f$ is unknown but is assumed to be Lipschitz-continuous.

Such problems arise notably in the field of machine learning\cite{Schulman2015,Bogolubsky2016,Salimans2017,Choromanski2018,Liu2018} including hyper-parameter tuning \cite{Snoek2012, BOHB, Hyperopt, Optuna} (see review \cite{Bischi2023}), automated architecture design\cite{Domhan2015} and reinforcement learning \cite{Gasnikov2022}. These problems also appear in the field of combinatorial optimization for algorithm tuning and configuration\cite{Hutter2007ParamILS,Hutter2009,Hutter2011SATTune,GGA,GGAp,f-race, Hoos2021, Bernal2021, Parizy2023} and automated design of search heuristics\cite{Khudabukhsh2009,Khudabukhsh2016}

In this paper, we are interested in cases where the evaluation of $\hat{f}$ is computationally very costly which puts a limit on the number points $x$ that can be sampled. As a case study, we focus more particularly on the task of algorithm tuning for heuristic NP-hard combinatorial optimization (CO) solvers. These heuristics\cite{SA, LKHeuristic, GSAT, Leleu2019, Leleu2021, Reifenstein2021, Reifenstein2023} have typically many hyper-parameters ($D \gg 1$) that need to be tuned for maximizing their performance\cite{Hutter2007ParamILS, Hutter2011SATTune, Hoos2021, Bernal2021, Parizy2023}. In this case, $\hat{f}$ can be interpreted as the objective function of the underlying CO problem evaluated by the heuristic solver configured with the parameters $x$ with $x \in \mathbb{R}^D$. The exact choice for the definition of $\hat{f}$ includes time to solution, probability of finding the optimal solution, decision variable of reaching or given solution quality or the solution quality itself. For applications such as quadratic unconstrained binary optimization (QUBO, equivalent to the Ising problem) or boolean satisfiability (SAT), the runtime of the heuristic typically scales exponentially with problem size $N$\cite{Karp2010}. Due to the random initialization of the initial state of the heuristic, the evaluation of $\hat{f}$ is very noisy. From a practical perspective, we also assume that it is possible to compute multiple evaluations of $\hat{f}(x)$ in parallel such as in a GPU.

It has been recognized that the parameter space in machine learning and combinatorial optimization exhibits heterogeneous (or anisotropic) curvature properties\cite{Sagun2016,Yao2020,Liu2023sophia,Leleu2021, Reifenstein2023} at large problem size. The anisotropic property implies that the distribution of eigenvalues $\{\lambda_i\}_{i \in \mathbb{R}^D}$ of the Hessian matrix $\nabla^2 f$ at proximity of its maximum ($\nabla f = 0$) possesses a heavy tail, i.e., some directions have a much higher curvature than the average with a high probability. While some standard machine learning methods such as ADAM\cite{ADAM} or second order methods\cite{Yao2021adahessian} attempt to adapt to the heterogeneous curvatures, the question of how to achieve accurate derivative-free optimization using curvature information in a noisy environment is not fully understood and is the subject of current research\cite{Bollapragada2019adaptive,Liu2020,Yao2021adahessian,Kunstner2023,Liu2023sophia}


\subsection{New Dynamical Smoothing Approach}

In this work, the methods of ball smoothing\cite{Gasnikov2022, Gasnikov2022R} and Gaussian smoothing\cite{Nesterov2017} for derivative-free optimization are extended to take into account the anisotropic curvature of the objective function $f$. In the new algorithm, the curvature of the smoothing kernel is dynamically adapted and converges to the Hessian matrix of the objective function $f$ at proximity of its local maxima. We prove that this dynamic sampling approach corrects for the anisotropy of the objective function which results in reduced error in the approximation of the gradient $\nabla_x f$ calculated from the noisy evaluations of $\hat{f}$ by the oracle. Numerical experiments on a set of artificial problems are conducted in order to demonstrate the benefits of this method. Moreover, we compare our algorithm against previously proposed derivative-free\cite{Gasnikov2022, Spall1998} and Bayesian optimization methods\cite{BOHB} on the task of parameter tuning for state-of-the-art NP-hard CO heuristic solvers and show that are approach is able to tune these heuristics more optimally by taking into account the different sensitivities of each parameter.

\subsection{Previous works}

The line of research that inspires this work is that of derivative-free optimization. In particular, these methods are known to work well in high dimensions just like regular gradient ascent/descent although an additional $O(D)$ cost is added \cite{Gasnikov2022}. In the presence of noise, it is not possible to get an accurate gradient using finite difference and many samples are needed to approximate the gradient. Gradient estimation can be done via finite differences\cite{Spall1998}, linear interpolation\cite{Conn2009introduction}, Gaussian smoothing\cite{Salimans2017,Nesterov2017}, and smoothing on a
unit sphere\cite{Gasnikov2022, Gasnikov2022R}. The convergence rate is given at best as $o(\frac{1}{\sqrt{n_s}})$ where $n_s$ is the number of samples\cite{Jamieson2012}. The trade-off between window size and measurement noise is investigated analytically in \cite{Gasnikov2022}. Derivative free optimization updating an approximate Hessian has been explored in \cite{Bollapragada2019}.

In addition to ball smoothing, there are many other methods that have been explored which aim to a function based on noisy measurements\cite{Spall1998,Kolda2003, Kim2010, Duchi2015,Conn2000, Deng2006, Sun2022}. Generally, these methods can be described as follows. The algorithm state is defined by some sort of sampling window which has both a position and a size. At each step, the algorithm makes noisy measurements of the objective function based on the sampling window. Then, based on the results of these samples we update the position and size of the window. Ideally, the window will shrink and the position will converge on the true optimum. The question of how to choose where to sample depending of the shape of $f$ is an open question.

There are many classes of approaches, each of which differs in exactly how the window is updated. Stochastic approximation (SA) methods such as \cite{Spall1998} approximate the gradient of $f$ with what is often called a stencil. In addition to SA, direct search (DS), also known as pattern search, uses similar ideas of moving around the parameter space and zooming in \cite{Kolda2003, Kim2010}. The main difference with pattern search is that typically the position is updated only when enough samples are taken to ensure the new position is better than the old one with high accuracy. This requires taking many samples in the same location. Trust region (TR) methods are another closely related method in which many samples are taken within a circular sampling window (trust region) and these samples are used to construct a model of the cost function\cite{Conn2000, Deng2006, Sun2022}. Based on this model we can estimate where the minimum of the objective function is within the trust region. We then evaluate the objective function to some accuracy at the perceived optimum. Depending on how this compares to the model we then update the window size and location. Like DS, this requires taking many samples of the objective function at the model optimum in order to be certain of the new window location and shape.

Early approaches that explore adapting the sampling window size include Nelder-Mead methods are a class of algorithms in which the algorithm state is defined by a simplex of $D+1$ points. Based on function evaluations of these points the size shape and position of the simplex is updated. Originally developed by Nelder and Mead \cite{Nelder1965} for noiseless functions this simplex method has been extended to noisy functions as well \cite{Barton1996}. Like DS and TR, this requires many function evaluations to ensure an accurate estimate.

One line of research which is closely related to this work is the tuning for hyper-parameters for neural networks in machine learning. Research in this area has often used techniques such as Bayesian optimization to efficiently find good parameter configurations for training neural networks. These methods work by first choosing a search region over which to look for good parameters and randomly sampling points in this region. Based on these samples of the objective function we can construct a model for the true objective function and, in turn, compute what is called an acquisition function. This acquisition informs us which point will be best to sample next based on estimating how much useful information we will get by evaluating the objective at that point. Bayesian optimization is very useful when we can only evaluate the objective function a few times due to the large computational cost, however it is known to struggle in larger dimensions ($D >= 20$) \cite{Frazier2018tutorial}. Recently many papers and libraries have used techniques from Bayesian optimization to construct algorithms designed for hyperparameter tuning. Some examples are BOHB (Bayesian optimization and hyperband) \cite{BOHB}, Optuna \cite{Optuna}, GPTune \cite{GPTune} among many others. A more complete review of previous works is included in appendix section \ref{sec: discussion}.




\section{Dynamic Window}

\begin{figure*}[t]
    \centering
    \includegraphics[width=0.25\textwidth]{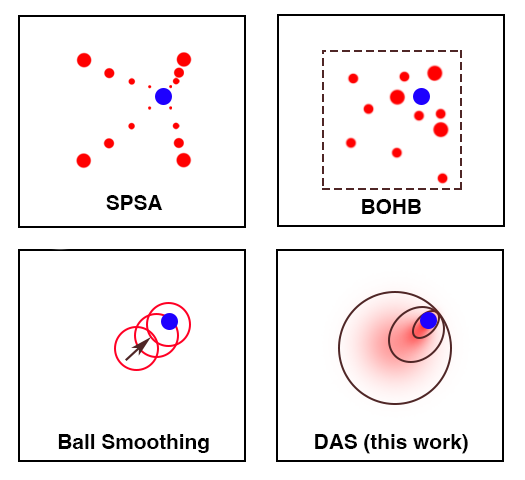}
    \includegraphics[width=0.33\textwidth]{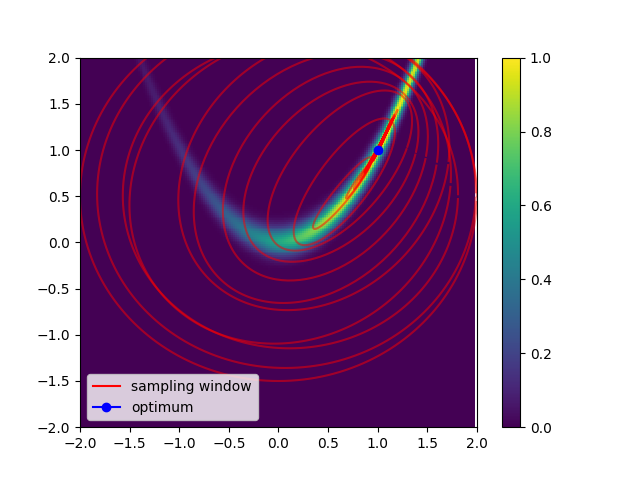}
    \includegraphics[width=0.33\textwidth]{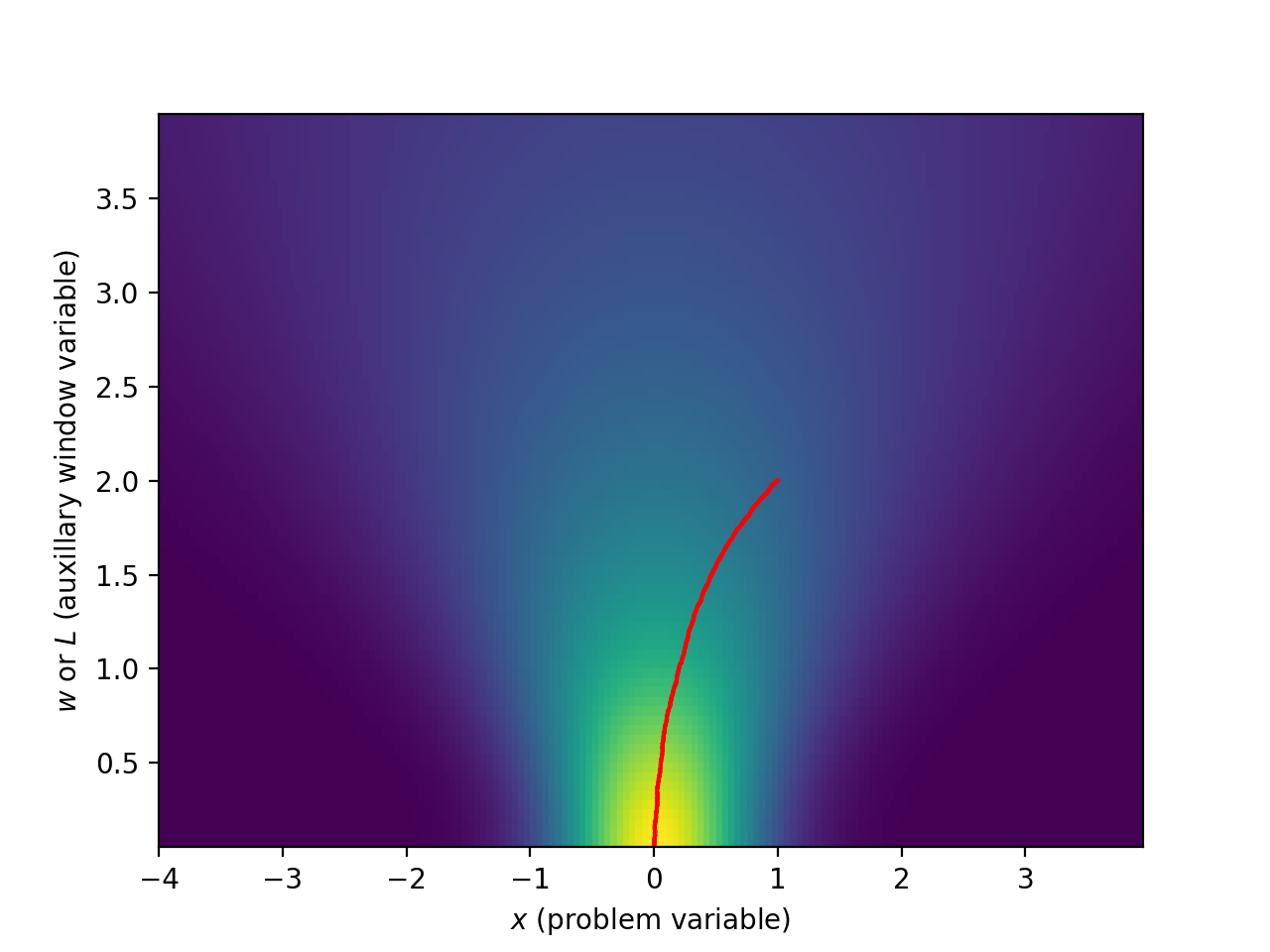}
   
    \caption{Left: Cartoon depiction of different derivative-free optimization methods discussed in this work while optimizing a 2-D objective function. Red represents the distribution of sampled points while the blue dot is the optimum. Middle: Sampling window of DAS while optimizing 2-dimensional modified Rosenbrock function (see section \ref{sec: rosenbrock} for details). Red ellipses represent one standard deviation of the Gaussian sampling distribution. As the algorithm progresses, the window adapts to the shape of the fitness function helping convergence. Right: Example of gradient ascent on $h(w,x)$ for a one-dimensional Gaussian fitness function in an ideal noiseless setting. The optimum is obtained by updating both window size $w$ and position $x$ according to the gradient of $h$ until the window size shrinks to 0 and $x$ converges to the optimum of $f$. This is the principle for which DIS (dynamic isotropic smoothing) and DAS (dynamic anisotropic smoothing) are based upon. }
     \label{fig: algorithm_overview_plots}
\end{figure*}

\subsection{Dynamic Isotropic Smoothing Algorithm (DIS)}

\label{sec: dynamic_ball_smoothing}
Before introducing the main algorithm studied in this work we will first present a simplified version which is an extension of ball smoothing \cite{Nesterov2017, Gasnikov2022, Gasnikov2022R}.
We will refer to this algorithm as dynamic isotropic smoothing (DIS). Similar to ball smoothing, this version of the algorithm considers gradient ascent on a smoothed version of $f$:
\begin{equation} \label{eq:smoothedf}
    h(w, x) = \frac{1}{w^D}\int \kappa( (u-x)/w) f(u)du,
\end{equation}

where the smoothing kernel $\kappa(x)$ we will define as a Gaussian $\kappa(x) = \frac{1}{(2\pi)^{D/2}}e^{-\sum_i x_i^2/2}$. 
\\
\\
We can then consider gradient ascent on $h$ with respect to the original coordinates $x$ but also an auxiliary coordinate $w$ which represents the window size. At each step the algorithm samples randomly according to the distribution $\kappa((u-x)/w)/w$. These samples are used to estimate $\nabla h$ and then a stochastic gradient descent (ascent) algorithm is used. This allows us to simultaneously update the window size ($w$) and position ($x$) following a straightforward mathematical formulation. In the right plot of Fig. \ref{fig: algorithm_overview_plots} we show how gradient ascent of $h$ converges on the optimum of $f$.


Next, we want to approximate $\nabla h(w,x)$ using the noisy samples of $f$ which we will denote $\hat{f}(x)$. To do this, we can first write $h_w(w,x) \in \mathbb{R}$ and $h_x (w,x) \in \mathbb{R}^D$ in the following integral forms.
\begin{flalign}
    h_x(w,x) = \int \frac{x-u }{w^{D+2}} \kappa( \frac{u-x}{w} ) f(u)du,
\end{flalign}
\begin{flalign}
    h_w(w,x) = \int \left( \frac{|x-u|_2^2}{w^{D+3}}- \frac{D}{w^{D+1}}\right) \kappa( \frac{u-x}{w} )  f(u)du.
\end{flalign}
These formulas express the gradient exactly, however, in practice, we only have noisy access to $f$ so clearly they cannot be computed exactly. 


When we sample $v$ according to some PDF
$P(v) = \frac{1}{w^D}\kappa((v-x)/w)$, then $h_x(w,x)$ and $ h_w(w,x)$ can be computed as follows:
\begin{flalign}
    \mathbb{E} \left( f(v)\frac{x-v}{w^{2}} \right) = h_x(w,x),
\end{flalign}
and
\begin{flalign}
    \mathbb{E} \left( f(v)\left(\frac{|x-u|_2^2}{w^{3}}- \frac{D}{w} \right)\right) = h_w(w,x)
\end{flalign}
These formulas allow for the gradients to be estimated by sampling the objective function so that gradient ascent dynamics can be implemented.

\subsection{Dynamic Anisotropic Smoothing Algorithm (DAS)}
\label{sec: dynamic_ellipsoid_smoothing}
In this section, we will extend the dynamics of the window size to also include window shape which we will call dynamic anisotropic smoothing (DAS). In other words, the window size can contract by different amounts in different directions to properly match the fitness function's shape. To do this, we create a new smoothed function this time parameterized by a matrix $L \in \mathbb{R}^D$ as:
\begin{flalign}
    h(L,x)  &= \int \kappa(v) f(Lv + x)dv  \\
    &= \text{det}(L)\int \kappa(L^{-1}(x-u))f(u)du
\end{flalign}

The algorithm is constructed by approximately simulating the following dynamics:
\begin{flalign} \label{eq: SDE}
    \frac{dL}{dt} = \alpha_L \left(LL^{\top} \frac{\partial h(L,x)}{\partial L} + \lambda L + \eta_L \right),  \\
    \frac{dx}{dt} = \alpha_x \left(LL^{\top} \frac{\partial h(L,x)}{\partial x} + \eta_x\right),
\end{flalign}

where $\lambda$ is a parameter controlling the growth of the window size with $\lambda>0$. For all numerical results in section \ref{sec: num_results} and \ref{sec: CO}, we use $\lambda = 0$. 

The $L L^{\top}$ factor is to keep the dynamics on the correct scale when the window changes size. $\alpha_L$ and $\alpha_x$ are chosen based on the properties of the fitness function. $\eta_L$ an $\eta_x$ are noise variables associated with the imprecise measurements of the gradients.

Similar to the previous section, we can compute the partial derivatives of $h$ with respect to $L$ and $x$. We then use gradient-ascent-like dynamics to devise an algorithm that accounts for heterogeneous curvature of the fitness function (to see why this is important refer to appendix \ref{sec: window_shape}).

The derivative of $h$ with respect to $L$ can be expressed as follows:
\begin{flalign}\label{eq: partial_L}
\frac{\partial h(L,x)}{\partial L} &= (L^{-1})^{\top}M, \\
M_{ij} &= -\int (v_i \kappa_j(v) + \delta _{ij} \kappa(v)) f(Lv + x)dv
\end{flalign}
and with respect to $x$ as:
\begin{flalign}\label{eq: partial_x}
    \frac{\partial h(L,x)}{\partial x} = -(L^{-1})^{\top}\int \nabla \kappa (v) f(Lv + x)dv.
\end{flalign}
When $\kappa$ is a Gaussian, these formulas can also be written as:
\begin{flalign}
    \frac{\partial h(L,x)}{\partial L} = (L^{-1})^{\top} \int (vv^{\top} - I) \kappa(v) f(Lv + x)dv,
\end{flalign}
\begin{flalign}
    \frac{\partial h(L,x)}{\partial x} = (L^{-1})^{\top} \int v \kappa(v) f(Lv + x)dv.
\end{flalign}

To simulate this SDE we approximate $\frac{\partial h(L,x)}{\partial L}$ and $\frac{\partial h(L,x)}{\partial x}$ using the following estimators:
\begin{flalign}
    \frac{\partial h(L,x)}{\partial L} &=  (L^{-1})^{\top}\mathbb{E}_{v, \zeta}(I + vv^{\top}\hat{f}(Lv + x, \zeta)), \label{eq: est_L} \\
    \frac{\partial h(L,x)}{\partial x} &= (L^{-1})^{\top}\mathbb{E}_{v, \zeta}(v\hat{f}(Lv + x, \zeta)).\label{eq: est_x}
\end{flalign}
Where $v$ is a Gaussian random vector and $\hat{f}(x, \zeta)$ is a noisy measurement of $f$ with random variable $\zeta$.
\\
\\
If we combine these concepts, we can use an Euler integration step along with the estimators to simulate the SDE (see eq. \eqref{eq: SDE}) which is described in the pseudocode shown in appendix \ref{sec: pseudo_code}. The parameters for this algorithm are $B_0$ (initial batch size), $\kappa$ (batch size exponent) and $\Delta t$ which is the integration time step as well as $\alpha_L$ and $\alpha_x$ in eq. \eqref{eq: SDE}. For all results in this paper, we use $\alpha_L = 1/D$, $\alpha_x = 1$ whereas the other parameters depend on the amount of parallelism that can be used and the properties of $f$. In this pseudo-code and in the numerical results in this work we schedule $B$ according to $\text{tr}(LL^{\top})^{\kappa/2}$ (size of the window) but other schedules can be used. The purpose of this is to use a more accurate gradient and when the window is smaller to resolve smaller and smaller values of $f$ accurately. This effect can also be achieved by manually setting a schedule for $B, \Delta t$, or both.
\\
\\
In the middle plot of Fig. \ref{fig: algorithm_overview_plots} we show the trajectory of the sampling window for DAS in the two-dimensional modified Rosenbrock function (defined by eq. (\ref{eq: mod_rosen}) in section \ref{sec: rosenbrock}). The sampling window starts large but as samples are collected it quickly shrinks around the nonzero region of the objective function. Because the nonzero region of the fitness function is constrained to a 1-D parabolic region, the sampling window shrinks more quickly along the axis perpendicular to the parabola. Once the window is small enough, it will begin to move along the parabola until it converges to the optimum.

\subsection{Fixed Points (with $\lambda \neq 0$)}
\label{sec: DAS_growth}

The fixed points of eq. (\Ref{eq: SDE}) can be expressed as follows:
\begin{align} \label{eq: fixed_points}
    -\lambda (LL^{\top})^{-1}_{ij} = \frac{\partial}{\partial x_i} \frac{\partial}{\partial x_j} h(x,L), \quad \nabla_x h(x,L) = 0.
\end{align}
In the limit that $\lambda \rightarrow 0$, $L \rightarrow 0$ and $x$ will approach a critical point of $f$. Moreover, $\lambda ^2 (LL^{\top})^{-1}$ will approach the hessian of $f$ at $x$. When $f$ is Gaussian, it can be shown that this fixed point corresponds to $-\lambda (LL^{\top})^{-1} = \nabla^2_x f(x)$ and $\nabla_x f(x) = 0$ even at large window size (see appendix section \ref{sec: fGaussian}). Given this interpretation for $L L^{\top}$, the gradient of eq. (\Ref{eq: SDE}) is multiplied by a pre-conditioner that converges to the inverse of the Hessian. As suggested by the literature on second-order optimization\cite{Boyd2004convex}, this type of pre-conditioner is optimal in the case of convex optimization.

\subsection{Error in Gradient Estimation}

The error in estimating the gradient of eq. (\ref{eq: est_x}) when the number of sampled points becomes large but finite is represented by the factor $\eta_x$ in eq. (\ref{eq: SDE}). It can be estimated at proximity of the maximum $\bar{x}$ of $f(x)$ and in the limit of a small window $|L| \rightarrow 0$. The central limit theorem implies that $\eta_x$ is normally distributed with a variance given as follows (assuming $f(\bar{x})$=1, see appendix section \ref{sec: variancedx}):

\begin{align}
\text{Var}[(\eta_x)_i]  &\approx \frac{1}{n_s} ( 
(L L^{\top})_{ii} f(\bar{x})^2 \nonumber \\
&+ \sum_{jkl} L_{ij}^2 (L^{\top} \nabla^2_x f(\bar{x}) L)_{kl}^2 \nu_{jkl} \nonumber \\
&- 2 f(\bar{x}) (L L^{\top})_{ii} \sum_{j} (L^{\top} \nabla^2_x f(\bar{x}) L)_{jj} \nonumber \\
&- 4 f(\bar{x}) (L (L^{\top} \nabla^2_x f(\bar{x}) L) L^{\top} )_{ii}
\end{align}

where $\nu_{jkl} = 2$ and $\nu_{jkl} = 15$ if exactly 2 and 3 subscript indices are equal, respectively;  $\nu_{jkl} = 1$ otherwise.

We show in appendix \ref{sec: variancedx} that the choice of $L$ that minimizes the gradient estimation error $E$ with $E = \text{Tr}[\{\text{Cov}[(\eta_x)_i,(\eta_x)_j]\}_{ij}] = \sum_i Var[(\eta_x)_i]$ must form an eigenbasis of the Hessian $H$ of $f$ near the maximum $\bar{x}$. This is obtained by finding $L$ that gives $\nabla_L E(L) \approx 0$ when also assuming that $L L^T$ has a constant Frobenius norm $|| L^T L ||_F$. In the limit of $|| L^T L ||_F \rightarrow 0$, the fixed point condition $\lambda (LL^T)^{-1} \approx \nabla_x^2 f$ of eq. (\ref{eq: fixed_points}) implies that $L$ can be expressed in the same eigenbasis as the Hessian of $f$ at its maximum. In other words, the DAS algorithm converges to the kernel curvature with smallest gradient estimation error.





\section{Numerical Results on Artificial Problems} \label{sec: num_results}

\subsection{Artificial Problems}

If the objective function can be measured with no noise or a sufficiently small amount of noise, there are derivative-free algorithms that can work with a fixed sampling window size \cite{Gasnikov2022, Gasnikov2022R}. That is, they estimate the gradient by measuring a few points that are close together. Because this allows for an accurate estimate of the gradient with $O(D)$ measurements, it can be shown that these methods only require $O(D)$ more samples than a regular gradient descent method \cite{Gasnikov2022, Nesterov2017}. In the presence of noise, this is no longer the case, however. In order to efficiently get an accurate measurement of the gradient we need to use a larger sampling window (or, in other words, the finite difference will be over a larger range). However, the gradient we get from this will not be that of the true objective function but a smoothed version of it. This biased gradient will cause the algorithm to smooth over finer details in the objective function. 
So, in the noisy case, there is a trade-off between large and small sampling windows which motivates the use of a dynamic sampling window which varies in size as the algorithm progresses. This trade-off is demonstrated numerically in appendix \ref{sec: tradeoff}. In appendix \ref{sec: additional_numres} we use artificial problems to demonstrate other properties of DIS and DAS such as motivation for changing window shape (appendix \ref{sec: window_shape}), and asymptotic scaling (appendix \ref{sec: asymptotic}).
\\
\\

\subsection{ Benchmark on Modified Rosenbrock Function}
\label{sec: rosenbrock}
\begin{figure*}[t]
    \centering
    \includegraphics[width=0.32\textwidth]{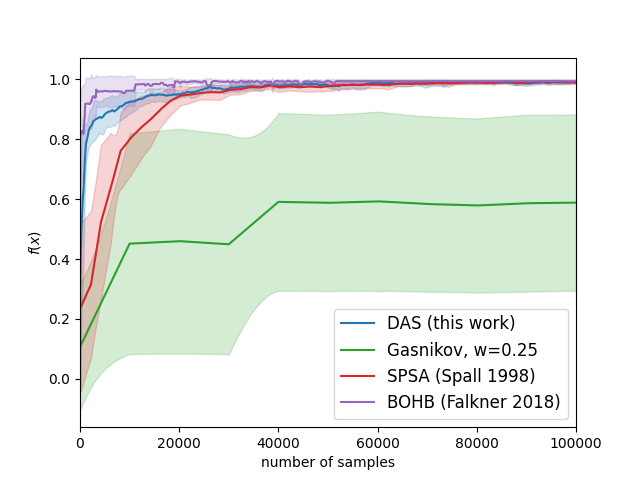}
    \includegraphics[width=0.32\textwidth]{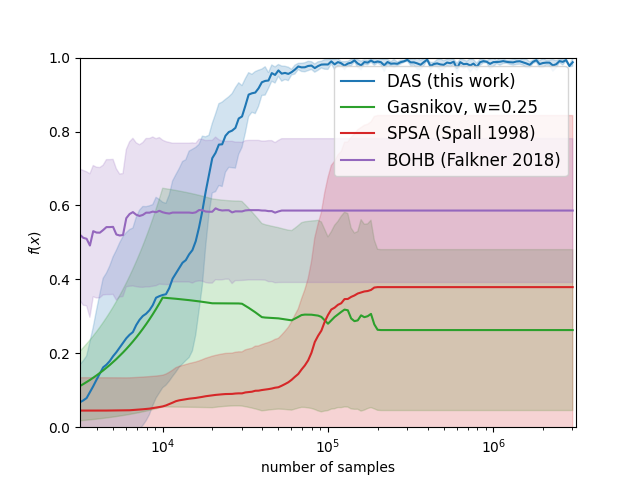}
    \includegraphics[width=0.32\textwidth]{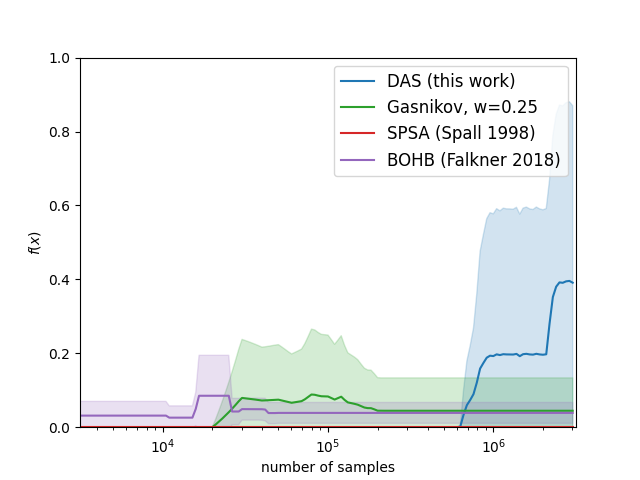}
    
    \caption{ Fitness as a function of $n_s$ on the modified Rosenbrock function in different dimensions for four different algorithms. Left: $D = 2, \beta = 0.5$, Middle: $D = 4, \beta = 0.5$, Right: $D = 8, \beta = 0.2$.  Traces are averages over 5 runs, and shaded regions represens one standard deviation of the data. Except for BOHB, each algorithm is initialized with an initial condition in $[0,1]^D$ and BOHB uses $[0,1]^D$ as the sampling window.}
    \label{fig: rosenbrock_bench}
\end{figure*}

A common test function used for derivative-free optimization algorithms in the past has been Rosenbrock Function defined as
\begin{equation}
    f(x) = \sum_{i=0}^{D-1} 100(x_{i+1} - x_{i}^2)^2 + (1-x_i)^2.
\end{equation}
In this work, we will mainly study a modified version of this function in which the fitness is restricted to be in the range $[0,1]$ similar to a probability.
\begin{equation}\label{eq: mod_rosen}
    f(x) = e^{-\beta \sum_{i=0}^{D-1} 100(x_{i+1} - x_{i}^2)^2 + (1-x_i)^2}
\end{equation}
This function exhibits many properties which make it hard for many algorithms to optimize. In particular, the first term (with a coefficient of 100) essentially restricts the optimal region to a parabolic manifold. Then, the optimizer will have to move along this manifold to optimize the second term which has a much more shallow gradient. Additionally, for each sample, we will randomly pick from $\{0,1\}$ with probability $f(x)$. This give the optimization problem similar characteristics to the CO parameter tuning discussed in section \ref{sec: CO}.
\\
\\
In Fig. \ref{fig: rosenbrock_bench} we show the result of applying four algorithms to this function in different dimensions and with different values of $\beta$. These numerical results are also summarized in table \ref{tab: rosenD4} and tables \ref{tab: rosenD2} and \ref{tab: rosenD8} of appendix \ref{sec: rosen_tables}. As expected, in lower dimensions ($D = 2$) BOHB almost always achieves the best fitness for any number of samples. However, when the dimension becomes larger, BOHB tends to struggle more and the three other methods, which are gradient based, are more effective. For the dimension $D=4$ case we can see that DAS has the best performance. Although SPSA is also effective in some cases, it can struggle from inconsistency with this type of objective function. That is, because of the discrete shape of the sampling region, many initial conditions will never be successful at all. In dimension $D=8$ all four algorithms struggle due to the large parameter space. However, with enough samples of the objective function, DAS is still able to find the optimum in many cases. For more detailed information about how DAS is implemented for the results in this section and section \ref{sec: CO}, see appendix \ref{sec: pseudo_code}.

\begin{table}
    \centering
    \begin{small}
    
    \begin{tabular} {c |p{0.07\textwidth} p{0.07\textwidth} p{0.07\textwidth}}
        \toprule
        &  mean &  worst &  best \\ \midrule
DAS (this work) &  \textbf{0.981} & \textbf{0.962} & \textbf{0.994} \\ 
Gasnikov, w=0.25  & 0.280 & 0.000 & 0.564 \\ 
SPSA (Spall 1998)  & 0.304 & 0.000 & 0.762 \\ 
BOHB (Falkner 2018)  & 0.586 & 0.243 & 0.796 \\
        \bottomrule
    \end{tabular} \label{tab: rosenD4}
    \caption{Table shows best, worst, and mean fitness achieved by four algorithms for $n_s = 10^{5}$. The toy function is the modified Rosenbrock function in 4 dimensions with $\beta = 0.5$.}
    
    \end{small}
\end{table}

\section{Application to combinatorial optimization \label{sec: CO}}

\subsection{Heuristic Solver Parameter Tuning}

Combinatorial optimization (CO) is a type of optimization in which we look for a solution that minimizes some objective function among a set of discrete configurations. The number of possible solutions increases exponentially with the problem size and, although the objective function can be computed in polynomial time, finding optimal configuration takes an exponential amount of time in the worst case for NP-hard problems. Examples include max-cut, max-clique, Ising, TSP (traveling salesman problem), the knapsack problem, graph coloring and SAT (boolean satisfiability).
\\
\\
Consequently, heuristic techniques have been developed that solve typical instances of many these problems efficiently with high probability but have no strict guarantees of finding the correct solution \cite{SA, LKHeuristic, GSAT}. More recently, a new class of heuristics, which we will call differential solvers, have been shown to be at least as efficient as state-of-the-art methods with the advantage of being well-fitted for implementation in specialized hardware accelerators\cite{ErcseyRavasz2011, Yamamoto2017, Leleu2019, Leleu2021, Reifenstein2021, Reifenstein2023, Goto2019, Goto2021, Kalinin2023}. These heuristic algorithms typically have many parameters that need to be tuned in order to work effectively (see appendix \ref{sec: co_disc} for more details). In this section, we will compare the effectiveness of state-of-the-art tuning methods at tuning the parameters of a recently developed differential SAT solver \cite{Reifenstein2023} (see appendix \ref{sec: co_details} for more details). Additionally, we have tuned a similar QUBO/Ising differential solver based on \cite{Leleu2019} and obtain similar results (see appendix \ref{sec: ising}).

\subsection{SAT Algorithm Tuning }

\begin{figure*}[t]
   
    \centering
    \includegraphics[width = 0.47 \textwidth]{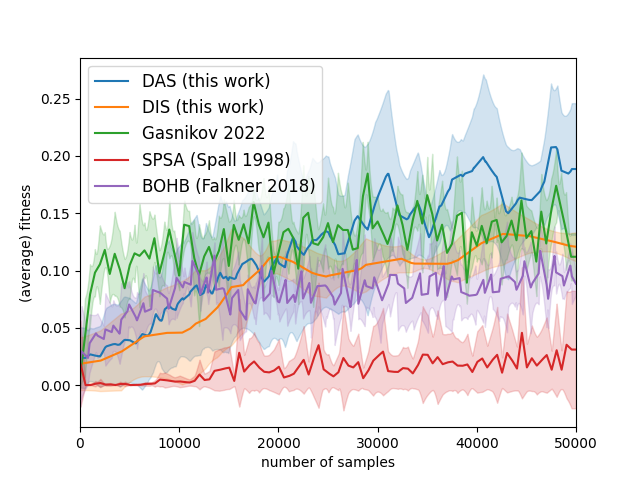}
    \includegraphics[width = 0.47 \textwidth]{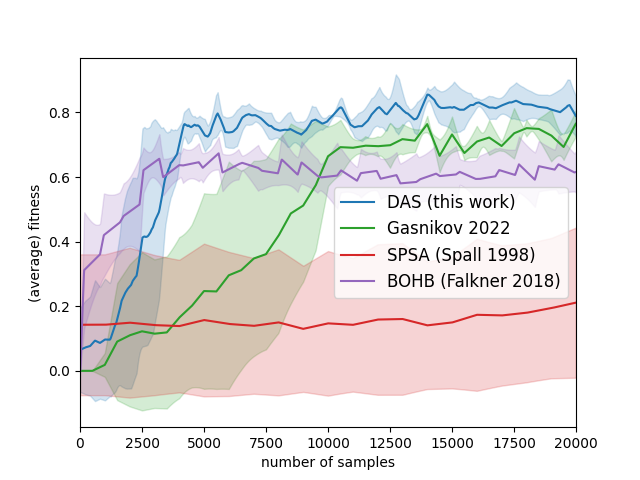}
    
    \caption{Left: Average success probability obtained by different tuning methods on random 3-SAT with $N = 150, \alpha = 4.0, T= 148$. Averages are over 5 realizations of the tuning dynamics starting at different randomized positions. The shaded region represents one standard deviation of the data. To evaluate the fitness of each parameter configuration, 20 random SAT instances are generated and 50 trajectories are evaluated for each. Right: A similar plot for tuning an Ising solver on problem size $N=150$ in which the performance improvement provided by DAS is clearer. This result is not discussed in the main text but is included in appendix \ref{sec: ising}}\label{fig: SAT_comparison}
    
\end{figure*}

In this section, we will consider the problem of tuning the SAT solver developed in \cite{Reifenstein2023} on random 3-SAT. This algorithm, which is described in more detail in appendix \ref{sec: co_details}, has four real parameters, $\text{dt}, p_{init}, p_{end}, \beta$. We want to find the values of these parameters which maximize the success rate for a certain class of SAT problems. 
More specifically, for a given problem size $N$, clause to variable ratio $\alpha$, and total number of time steps $T$ there is a function of four real variables $P_{avg}(\text{dt}, p_{init}, p_{end}, \beta) \in [0,1]$  which we wish to optimize. The noisy evaluation of this function is realized by choosing a random 3-SAT problem with the relevant parameters ($N$, $\alpha$) and computing a single trajectory of the coherent SAT solver with the given number of time steps $T$ and the given system parameters. Thus, the problem of choosing optimal parameters is a noisy derivative-free optimization problem.
\\
\\
In Fig. \ref{fig: SAT_comparison} we compare the performance of different optimizers discussed in this paper on tuning the SAT solver with $N = 150$, $\alpha = 4.0$ $T = 148$. Our proposed algorithm, DAS, is able to find the best parameters. This is mostly because, unlike the other methods, it can properly account for the different sensitivities of the different parameters (see appendix \ref{sec: tune_traj} for more details). One important detail to touch upon for the results in Figs. \ref{fig: SAT_comparison} (and Fig. \ref{fig: SAT_tune_trajectories} of appendix section \ref{sec: tune_traj}) is the optimal parameters found by DAS have very extremal values, that is, $p_{init} \approx -1$ and $\beta \approx 2$. In fact, these values are out of the search space given to BOHB which is $[0,1]^{4}$. In some sense, this is unfair to BOHB because it cannot find parameters outside of the bounds of its search. However, this is precisely one of the reasons that we believe BOHB is sub-optimal for this type of problem. Expanding the search space of BOHB to $[-1,2]^4$ would encompass the optimal parameters, but this adjustment would exponentially increase the search area by a factor of $3^4$ in the worst case, potentially leading to significant slowing down of the parameter tuning. While it is tempting to narrow the first dimension of the search box to $[0, 0.3]$, based on the knowledge that the critical parameter $dt$ is optimally near $0.1$, such a strategy would inherently rely on specific prior insights. For the purpose of this discussion, we operate under the premise of minimal prior knowledge regarding the parameters, aside from a basic understanding of their magnitude. This constraint is crucial for an unbiased evaluation of a tuning method's effectiveness, enabling its application across diverse algorithms and parameter sets without presupposed knowledge, thereby allowing for an autonomous determination of parameter sensitivities. This approach underscores the adaptability and effectiveness of the DAS method in tuning scenarios. To further emphasize the improvement provided by DAS, Fig. \ref{fig: SAT_comparison} also includes results on tuning an Ising solver. These results are discussed in more detail in appendix \ref{sec: ising}.

\section{Conclusion}

In this work we have presented a new algorithm, which uses previous ideas from derivative-free optimization\cite{Gasnikov2022,Gasnikov2022R,Nesterov2017}. We have shown that it is an effective tool for the tuning of combinatorial optimization heuristic solvers. Due to its ability to adapt to a noisy objective function with heterogeneous curvature, DAS can outperform existing methods of parameter tuning\cite{Gasnikov2022,Spall1998,BOHB} for a variety of applications\cite{Leleu2021, Reifenstein2023}. The pros and cons of these different methods are briefly summarized in table \ref{tab: review}. The advantage of the method for dealing with heterogenous curvature is justified analytically by the calculation of the reduction of gradient estimation error induced by the dynamic anisotropic kernel. Based on our experimentation with artificial objective functions, we believe that key properties that our target application has are: 1) heterogeneous curvature (meaning different sensitivities of parameters and correlations between parameters), 2) an objective function that evaluates to zero in most cases but is nonzero with some probability in the region of interest. We believe that in addition to combinatorial optimization, there are many other applications in which the objective function has these properties (such as the training of neural networks)\cite{Sagun2016,Yao2020,Liu2023sophia}. Thus, our new method likely has many uses beyond what is discussed in this work.
\\
\\
There are also many ways in which this method can be extended to improve its effectiveness. The approach builds upon prior works by allowing additional degrees of freedom in the sampling process. Future studies could explore extensions like incorporating higher-order moments, using sums of Gaussians for the sampling window, adding momentum to gradient ascent for enhanced performance\cite{ADAM}, and employing preconditioners for a stable Hessian estimate, particularly beneficial in neural network configuration (see discussion in appendix \ref{sec: hessian_ML_rev}). Further, the use of common random numbers in objective function evaluations, a technique proven to boost the efficacy of derivative-free optimization methods\cite{Agarwal2010, Shamir2015}, presents another promising avenue for improvement.

\begin{table*}[b]
\begin{center}
    
\begin{small}
    
    \begin{tabular}{|p{0.12\textwidth}||p{0.19\textwidth} |p{0.19\textwidth}|p{0.19\textwidth}|p{0.19\textwidth}|}
    \toprule
       Type of Method & Bayesian Optimization & Fixed Sampling Window & Dynamic Window Size & Dynamic Window Size and Shape \\
       \hline \hline
       References & BOHB \cite{BOHB}, HyperOpt \cite{Hyperopt}, Optuna \cite{Optuna}, GPTune \cite{GPTune}, (review \cite{Bischi2023}) & Gradient Estimator Methods \cite{Nesterov2017, Gasnikov2022}, (review \cite{Gasnikov2022R,Berahas2022})  & SA (stochastic approximation) \cite{Robbins1951, Spall1998}, DS (direct search) \cite{Kolda2003, Kim2010}, TR (trust region methods) \cite{Conn2000, Deng2006, Sun2022}, (review \cite{Larson2019}) & Modified Nelder-Mead methods \cite{Nelder1965, Barton1996}, DAS (This Work) \\ \hline
       Pros & Can handle noisy evaluation of non-convex function & Good for high dimensions, simple & Handles noisy function evaluations better & Can handle heterogeneous curvature of fitness function \\ \hline
       Cons & Struggles with high dimensions & Struggles to deal with noise & Struggles with heterogeneous curvature of fitness function &  Restricted to continuous search space\\ \hline
    \end{tabular}
    
\end{small}
\end{center}
    \caption{Table comparing autotuning methods described in this paper.}
    \label{tab: review}
\end{table*}




\section*{Software and Data}

\section*{Acknowledgements}


\newpage
\appendix
\onecolumn

\section{Derivation of Analytical Results}

\subsection{Derivation of Equations \eqref{eq: partial_L} and \eqref{eq: partial_x}}

Equations \eqref{eq: partial_L} and \eqref{eq: partial_x} can be derived as follows.
$$  h(L,x) = \int \kappa(v) f(Lv + x)dv \quad \rightarrow \left(\frac{\partial h(L,x)}{\partial L}\right)_{ij} = \int \kappa(v) \frac{\partial}{\partial L_{ij}} f(Lv + x)dv$$
$$ =  \int \kappa(v) v_j f_i(Lv + x)dv$$
This integral uses partial derivatives of $f$ which we do not have oracle access to, however, we can use integration by parts to get rid of this. First we need to substitute $u =  Lv$, $du = \text{det}(L)dv$.
$$ \int \kappa(v) v_j f_i(Lv + x)dv = \int \kappa(L^{-1}u)\left( \sum_k (L^{-1})_{jk}u_k \right) f_i(u + x)\text{det}(L)^{-1} du $$
Then we can use integration by parts on the $u_i$ coordinate to get:
$$= -\int \left( \frac{ \partial \kappa(L^{-1}u)}{\partial u_i}\left( \sum_k (L^{-1})_{jk}u_k \right) +  \kappa(L^{-1}u)(L^{-1})_{ji}   \right)  f(u + x)\text{det}(L)^{-1} du $$
$$ = -\int \left( \left( \sum_k \kappa _k(L^{-1}u) (L^{-1})_{ki} \right) \left( \sum_k (L^{-1})_{jk}u_k \right) +  \kappa(L^{-1}u)(L^{-1})_{ji}   \right)  f(u + x)\text{det}(L)^{-1} du$$
Re substituting $u =  Lv$ we get:
$$ = -\int  \left( \left( \sum_k \kappa _k(v) (L^{-1})_{ki} \right)v_j +  \kappa(v)(L^{-1})_{ji}   \right) f(Lv + x) dv $$
$$ = - \sum_k (L^{-1})_{ki} \int  \left( \kappa _k(v)  v_j +  \kappa(v)\delta_{jk}   \right) f(Lv + x) dv $$

\subsection{Explanation of Equation \eqref{eq: SDE}: Scale Symmetry} \label{sec: SDE_exp_1} 

Equation \eqref{eq: SDE}, which algorithm \ref{alg: SDE_tuner} is based upon, can be interpreted as a simple stochastic gradient ascent equation except for the extra term $LL^{\top}$ which multiplies both gradients. In this section, we will give explanations for this extra factor.
\\
\\
Since $LL^{\top}$ is closely related to the hessian of the fitness function (see eq. (\ref{eq: fixed_points}), one interpretation is that this matrix is a sort of preconditioner that is used to improve the numerical stability of the algorithm. It is known that using the inverse of the hessian as a preconditioner allows for much faster convergence of gradient descent-based methods. However, $LL^{\top}$ is not exactly the inverse of the hessian especially when no growth term is included.
\\
\\
A more concrete and complete explanation is that this factor ensures that the dynamics exhibit a sort of scale symmetry as follows. If we are given an objective function $f$ and an arbitrary nonsingular matrix $M$ we can construct the objective function $f_M(x)$ defined by $f_M(x) = f(Mx)$. Then we consider equation \eqref{eq: SDE} without noise for now. That is:

\begin{equation} \label{eq: ODE}
    \frac{dL}{dt} = \alpha_L  LL^{\top} \frac{\partial h(L,x)}{\partial L}, \quad \frac{dx}{dt} = \alpha_x LL^{\top} \frac{\partial h(L,x)}{\partial x},
\end{equation}
and then consider the variable substitution $ML_M= L$, $Mx_M = x$ which gives us:
$$M\frac{dL_M}{dt} = \alpha_L \left(M L_M L_M^{\top}(M)^{\top} \frac{\partial h(ML,Mx)}{\partial L} \right)\quad M\frac{dx_M}{dt} = \alpha_x M L_M L_M^{\top}(M)^{\top} \frac{\partial h(ML_M,Mx_M)}{\partial x}
$$
which simplifies to
\begin{equation} 
    \frac{dL_M}{dt} = \alpha_L L_ML_M^{\top} \frac{\partial h_M(L_M,x_M)}{\partial L_M}, \quad \frac{dx_M}{dt} = \alpha_x L_ML_M^{\top} \frac{\partial h(L_M,x_M)}{\partial x_M}.
\end{equation}
This tells us that the dynamics are invariant under arbitrary linear transformations which is a nice property to have. In other words, no matter how squeezed and shrunk the sampling window gets, the dynamics are in some sense identical to the original dynamics with a circular window.

\subsection{Derivation of Equation \eqref{eq: fixed_points}}

Equation \eqref{eq: fixed_points} can be derived as follows. First we will consider equation \eqref{eq: SDE} with $\frac{d L}{dt} = 0$ and $\frac{dx}{dt} = 0$. This gives us
$$L L^{\top} \frac{\partial h(L,x)}{\partial L} + \lambda L = 0$$
and
$$ \nabla_x h(x,L) = 0 $$
respectively. For the first equation, we can rewrite it by considering another expression for $ \frac{\partial h(L,x)}{\partial L} $ as follows.
$$  h(L,x) = \int \kappa(v) f(Lv + x)dv \quad \rightarrow \left(\frac{\partial h(L,x)}{\partial L}\right)_{ij} = \int \kappa(v) \frac{\partial}{\partial L_{ij}} f(Lv + x)$$
$$\int \kappa(v) \frac{\partial}{\partial L_{ij}} f(Lv + x) = \int \kappa(v)v_j f_i(Lv + x)dv = -\int \kappa_j(v) f_i(Lv + x)dv $$
(because $\kappa$ is a Gaussian, $- v_j \kappa(v) = \kappa_j(v)$) Then we can use integration by parts on the $v_j$ coordinate to get:
$$-\int \kappa(v) \frac{\partial}{\partial v_j}f_i(Lv + x)dv = \int \kappa(v) \sum_{k} L_{kj}f_{ki}(Lv + x)dv $$
$$\sum_{k} L_{kj}\frac{\partial}{\partial x_k}\frac{\partial}{\partial x_i} h(L,x) = \left(\frac{\partial h(L,x)}{\partial L}\right)_{ij} $$
$$ HL = \frac{\partial h(L,x)}{\partial L} $$
Where $H_{ij} =  \frac{\partial}{\partial x_i} \frac{\partial}{\partial x_j} h(x,L)$
So
$$LL^{\top}HL + \lambda L = 0$$
$$LL^{\top}H = -\lambda$$

\subsection{Derivation error in gradient estimation \label{sec: variancedx}}


In the following, we compute the error in the estimation of the gradient when the number of sampled points become large but finite. The gradient $dx = L L^{\top} \frac{\partial h(L,x)}{\partial x}$ can be written as $dx = L L^{\top} \frac{\partial h(L,x)}{\partial x}  = L  \frac{1}{n_s} \sum_{v \in V} (v\hat{f}(Lv + x, \zeta))$ where $n_s$ is the number of samples and $dx$ is a sum over of many independent and identically distributed variables $v$. We set $\alpha_x = 1$ and $dt = 1$ for the sake of simplicity.

More particularly, we are interested in finding the matrix $L$ that minimizes the total error $E$ defined as follows:

$$E = \text{Tr}[\{\text{Cov}[dx_i,dx_j]\}_{ij}].$$

That is to say, we are looking for $L$ such that $\nabla_L E = 0$. Importantly, we also assume that $L L^T$ has a constant Frobenius norm $|| L^T L ||_F$, i.e., $\text{Tr}[L L^T] = S$, $\forall L$ with $S>0$.
 
The displacement $dx$ can be simplified at proximity of the maximum $\bar{x}$ of $f(x)$ and in the limit of a small window $|L| \rightarrow 0$.

First, we have:

$$f(Lv + \bar{x}) = f(\bar{x}) + Lv \nabla_x f(\bar{x}) + (Lv)^T \nabla^2_x f(\bar{x}) (Lv) + o(S^2),$$

with $ \nabla_x f(\bar{x}) = 0$. The displacement $dx$ can be approximated as follows:

$$dx  = \frac{1}{n_s} \sum_{v} [f(\bar{x}) X(v) + Y(v) + o(S^2)],$$

with

\begin{align}
     X(v) &= L v,\\
     Y(v) &= L v [v^{\top} (L^{\top} \nabla^2_{x} f(\bar{x}) L) v ].
\end{align}

The expression for $\text{Var}[(dx)_i] = \text{Var}[(\eta_x)_i]$ is:

$$\text{Var}[(\eta_x)_i]  = \frac{1}{n_s} [f(\bar{x})^2 \text{Var}[X_i(v)] + \text{Var}[Y_i(v)] - 2 f(\bar{x}) \text{Cov}[X_i(v),Y_i(v)] +  o(S^2)],$$

where $\text{Var}[X_i(v)]$, $\text{Var}[Y_i(v)]$, and $\text{Cov}[X_i(v),Y_i(v)]$ can be expressed as follows:

\begin{align}
     (\text{Var}[X_i(v)] &= \sum_j L_{ij}^2 f(\bar{x})^2,\\
     (\text{Var}[Y_i(v)]) &= \sum_{jkl} L_{ij}^2 (L^{\top} \nabla^2_x f(\bar{x}) L)_{kl}^2 \nu_{jkl},\\
     (\text{Cov}[X_i(v),Y_i(v)]) &= \sum_{jk} L_{ik}^2 (L^{\top} \nabla^2_x f(\bar{x}) L)_{jj} + 2 \sum_{jk} L_{ij} L_{ik} (L^{\top} \nabla^2_x f(\bar{x}) L)_{jk}.
\end{align}

where $\nu_{jkl} = 2$ and $\nu_{jkl} = 15$ if exactly 2 and 3 subscript indices are equal, respectively;  $\nu_{jkl} = 1$ otherwise (for e.g., $\nu_{jjj} = 15$, $\forall j$, and $\nu_{jll} = 2$, $\forall j \neq l$). Thus, we have:


\begin{align}
\text{Var}[(\eta_x)_i]  &= \frac{1}{n_s} ( \sum_j L_{ij}^2 f(\bar{x})^2,\nonumber \\
&+ \sum_{jkl} L_{ij}^2 (L^{\top} \nabla^2_x f(\bar{x}) L)_{kl}^2 \nu_{jkl}, \nonumber \\
&- 2 f(\bar{x}) \sum_{jk} L_{ik}^2 (L^{\top} \nabla^2_x f(\bar{x}) L)_{jj}, \nonumber \\
& - 4 f(\bar{x}) \sum_{jk} L_{ij} L_{ik} (L^{\top} \nabla^2_x f(\bar{x}) L)_{jk} + o(S^3) )),
\end{align}

which can be written as

\begin{flalign}
\text{Var}[(\eta_x)_i]  &= \frac{1}{n_s} ( 
(L L^{\top})_{ii} f(\bar{x})^2 \nonumber \\
&+ \sum_{jkl} L_{ij}^2 (L^{\top} \nabla^2_x f(\bar{x}) L)_{kl}^2 \nu_{jkl} \nonumber \\
&- 2 f(\bar{x}) (L L^{\top})_{ii} \sum_{j} (L^{\top} \nabla^2_x f(\bar{x}) L)_{jj} \nonumber \\
&- 4 f(\bar{x}) (L (L^{\top} \nabla^2_x f(\bar{x}) L) L^{\top} )_{ii} + o(S^3) )),
\end{flalign}

Consequently, we have with $H = \nabla^2_x f(\bar{x})$ and $M = L^{\top} H L$:

\begin{flalign}
\sum_i \text{Var}[(\eta_x)_i] &= \frac{1}{n_s} ( \text{Tr}[L L^{\top}] f(\bar{x})^2 \nonumber \\ 
&+ \text{Tr}[L L^{\top}] \text{Tr}[(L^{\top} H L)^2] + \text{Tr}[J (L^{(2)} \circ M^{(2)})]\nonumber \\ 
&- 2 f(\bar{x}) \text{Tr}[L L^{\top}] \text{Tr}[L^{\top} H L] \nonumber \\ 
&- 4 f(\bar{x}) \text{Tr}[L (L^{\top} H L) L^{\top}] + o(S^3)), 
\end{flalign}

where $J$ is square matrix of ones, $L^{(2)}$ and $M^{(2)}$ element-wise square matrices $L$ and $M$, respectively.

We get using the approximation $\text{Tr}[ (L^{\top} H L)^2 ] \approx \text{Tr}[ L^{\top} H L ]$ and $\text{Tr}[ L (L^{\top} H L) L^{\top} ] \approx \text{Tr}[ L^{\top} L ] \text{Tr}[ L^{\top} H L ]$):

\begin{flalign}
\sum_i \text{Var}[(\eta_x)_i] &= \frac{1}{n_s} ( \text{Tr}[L L^{\top}] f(\bar{x})^2 \nonumber \\ 
&+ \text{Tr}[L L^{\top}] \text{Tr}[ L^{\top} H L ]^2\nonumber \\ 
&- 2 f(\bar{x}) \text{Tr}[L L^{\top}] \text{Tr}[L^{\top} H L] \nonumber \\ 
&- 4 f(\bar{x}) \text{Tr}[ L^{\top} L ] \text{Tr}[ L^{\top} H L ] + o(S^3)).\label{eq: approx_eta_x} 
\end{flalign}

We want to know how $E = \sum_i \text{Var}[(\eta_x)_i]$ changes after an infinitesimal change in $L$ under the constraint that the Frobenius norm $||L^{\top} L||_F^2$ remains constant, i.e., $\text{Tr}[L L^{\top}] = S$. For this, we define the following Lagrangian:

$$\mathcal{L}(L,\lambda) = E(L) + \mu (\text{Tr}[L L^{\top}] - S).$$ 

where $\mu$ is a scalar Lagrangian multiplier. To find $L$, we compute the gradient of $\mathcal{L}$ given as follows:

\begin{flalign}
\nabla_L \mathcal{L} &= \nabla_L E(L) + 2 \mu L.
\end{flalign}

Note that:

\begin{flalign}
\nabla_L \text{Tr}[L L^{\top}] &= 2L,\\
\nabla_L \text{Tr}[L^{\top} H L] &= 2HL,
\end{flalign}

It implies that:

\begin{flalign}
\nabla_L E[L] &= \frac{1}{n_s} ( 2L f(\bar{x})^2 \nonumber \\ 
&+ 2L \text{Tr}[ (L^{\top} H L) ]^2 + 4HL \text{Tr}[L L^{\top}] \text{Tr}[ (L^{\top} H L) ]\nonumber \\ 
&- 2 f(\bar{x}) (2L \text{Tr}[L^{\top} H L] + 2HL \text{Tr}[L L^{\top}]) \nonumber \\ 
&- 4 f(\bar{x}) (2L \text{Tr}[ L^{\top} H L ] + 2 H L\text{Tr}[ L^{\top} L ] + o(S^3) ). 
\end{flalign}

Then, $\nabla_L \mathcal{L} = 0$ and $\nabla_{\mu} \mathcal{L} = 0$ implies:

\begin{align}
0 &= \nabla_L E(L) + 2 \mu L,\\
0 &= 2L f(\bar{x})^2 + 2L \text{Tr}[ L^T H L ]^2 + 4HL \text{Tr}[L L^{\top}] \text{Tr}[L^T H L] \nonumber\\
&- 2 f(\bar{x}) (2L \text{Tr}[L^T H L] + 2HL \text{Tr}[L L^T]) - 4 f(\bar{x}) ( 2L \text{Tr}[ L^T H L ] + 2 H L\text{Tr}[ L^T L ] ) + 2 \mu n_s L + o(S^3),\nonumber\\
0 &= 2L f(\bar{x})^2 + 2L \text{Tr}[ L^T H L ]^2 + 4HL S \text{Tr}[L^T H L] \nonumber\\
&- 2 f(\bar{x}) (2L \text{Tr}[L^T H L] + 2HL S - 4 f(\bar{x}) ( 2L \text{Tr}[ L^T H L ] + 2 H LS ) + 2 \mu n_s L + o(S^3),\\
\end{align}

Lastly, it is possible to rewrite the condition of the solutions of $\nabla_L \mathcal{L} = 0$ and $\nabla_{\mu} \mathcal{L} = 0$ as follows:
\begin{align}
HL \approx \gamma' L,
\end{align}
with $\gamma'$ a scalar. i.e., $L$ must align with an eigenvector of the Hessian $H$ of $f$ near the maximum $\bar{x}$ (i.e., one column of $L$ is an eigenvector). More generally, $L$ aligns with all eigenvectors of forming the basis of H.



When $-\lambda(LL^T)^{-1} = H$ at the fixed point of the dynamics of $L$ and $x$ (see eq. (\ref{eq: fixed_points})), $L$ and $H$ align to the same set of eigenvectors. In this section, we have shown that the gradient estimation error $\sum_i \text{Var}[(\eta_x)_i]$ is minimized when $L$ and $H$ share the set of eigenvectors. Thus, gradient estimation error is minimized at the fixed point.

\begin{figure}
    \centering
    \includegraphics[width=0.5\textwidth]{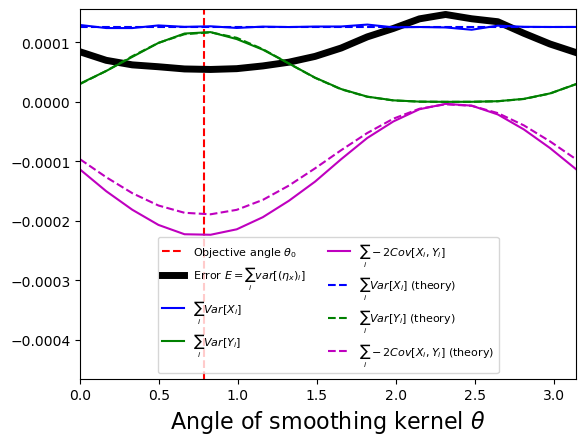}
    
    \caption{Error in gradient estimation $E = \sum_i Var[(\eta_x)_i]$ of the gradient of $x$ with $\text{Var}[(\eta_x)_i] = (\text{Var}[X_i] + \text{Var}[X_i] - 2 \text{Cov}[X_i,Y_i] )$ for a Gaussian objective function $f(x) = \kappa(M(x-\bar{x}))$ centered in $\bar{x}$ and Hessian $H$ with eigenvalues $\lambda'_1 = 1$ and $\lambda'_2 = 4$ rotated by a reference angle $\theta_0$ and anisotropic smoothing kernel centered in $\bar{x}$ and curvature $(L L^{\top})^{-1}$ with  with eigenvalues $\lambda_1 = 0.01$ and $\lambda_2 = 0.04$ and rotation $\theta$. The gradient estimation error $E$ is minimized for $\theta = \theta_0$. The dashed lines show the approximations of eq. (\ref{eq: approx_eta_x}). $D=2$. \label{fig: variancedx}} 
\end{figure}

Figure \Ref{fig: variancedx} shows that the covariance $-{\text{Cov}[X_i(v),Y_i(v)]}_i$ and, in turn, the variance of $dx$ is minimized when the smoothing kernel is aligned to the curvature of the objective function. That is, the noise in the step $dx$ is minimized when the window shape matches the curvature of $f$ near its maximum.


\subsection{Case of $f$ Gaussian with heterogeneous curvature \label{sec: fGaussian}}

First note that $x$ reaches the maximum of $f$ (i.e., $\nabla_x h(x,L)=0)$ when $f$ is convex under the dynamics of eq. (\ref{eq: SDE}). This is because

\begin{equation} \label{eq: convex_dhdx}
\frac{\partial h(L,x)}{\partial x}  = 0 \implies \frac{\partial f(x)}{\partial x} = 0,
\end{equation}

if $f$ is convex (i.e., $\frac{\partial f(x)}{\partial x}$ does not change sign) with

\begin{equation}
\frac{\partial h(L,x)}{\partial x} = \int \kappa(v) \frac{\partial f(Lv + x)}{\partial x} dv,
\end{equation}

and $\kappa(v)>0$, $\forall v$.

We consider the simpler case when $f$ is a Gaussian function. In this case, we show that the dynamic window $LL^T$ with growth term converges to the Hessian of $f$ at its maximum (i.e., inverse of covariance matrix).

When $f$ is a Gaussian function with center $\bar{x}$ and inverse covariance matrix (Hessian) $M M^T$, i.e., $f(x) = \kappa(M(x-\bar{x}))$, the fixed point of eqs. (\ref{eq: fixed_points}) can be written as follows:

\begin{equation} \label{eq: fixed_points_Gaussian}
    -\lambda (LL^{\top})^{-1}_{ij} = (MM^{\top})_{ij}, x = \bar{x}.
\end{equation}

To show eqs. (\ref{eq: fixed_points_Gaussian}), first note that $\frac{\partial h(L,x)}{\partial x}  = 0 \implies \frac{\partial f(x)}{\partial x} = 0$ when $f$ is Gaussian using eq. (\ref{eq: convex_dhdx}) and, therefore, $x = \bar{x}$ at the fixed point. Moreover, $\frac{\partial}{\partial x_i} \frac{\partial}{\partial x_j} h(L,x) = (MM^{\top})_{ij}$ at $x = \bar{x}$ because

$$  \frac{\partial}{\partial x_i} \frac{\partial}{\partial x_j} h(L,x) = \int \kappa(v) \frac{\partial}{\partial x_i} \frac{\partial}{\partial x_j} \kappa(M(Lv + x-\bar{x})) dv,$$

with $$ \frac{\partial}{\partial x_i} \frac{\partial}{\partial x_j} \kappa(M(Lv + x-\bar{x})) = 4 (MM^{\top} (Lv + x - \bar{x}))_{i} (MM^{\top} (Lv + x - \bar{x}))_{j} \kappa(M(Lv + x-\bar{x})) - 2 (MM^{\top})_{ij} \kappa(M(Lv + x-\bar{x})).$$

At $x=\bar{x}$,  we have 

$$ \int \kappa(v) (MM^{\top} (Lv))_{i} (MM^{\top} (Lv))_{j} \kappa(M(Lv)) dv = 0, $$

and

$$ \int \kappa(v) \kappa(M(Lv)) dv = 1. $$

\begin{figure}
    \centering
    \includegraphics[width=0.4\textwidth]{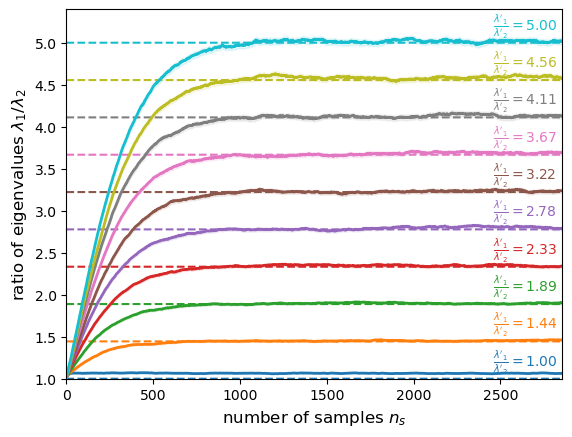}
    
    \caption{Ratio $\frac{\lambda_1}{\lambda_2}$ of the eigenvalues of the Hessian of the smoothing kernel $(LL^{\top})^{-1}$ noted $\lambda_1$ and $\lambda_2$ vs. the number of samples for various ratio $\frac{\lambda'_1}{\lambda'_2}$ of the Hessian of $f$ when $f$ is a Gaussian function. The dynamics converge to $\frac{\lambda_1}{\lambda_2} = \frac{\lambda'_1}{\lambda'_2}$. $D=2$.}
    \label{fig: convergenceLverified}
\end{figure}

Consequently, $\frac{\partial}{\partial x_i} \frac{\partial}{\partial x_j} h(L,x) = (MM^{\top})_{ij}$. Note that this property is true whatever the window size when $f$ is Gaussian, but is not necessarily true for a general function $f$. Numerical simulations shown in Fig. \ref{fig: convergenceLverified} confirm that the Hessian of the Gaussian smoothing kernel converges to that of the function $f$ in the case $d=2$.


\clearpage

\section{Algorithm Pseudo-Code and Numerical Stability} \label{sec: pseudo_code}


\begin{algorithm}[h]
   \caption{DAS (Dynamic Anisotropic Smoothing)}
\begin{algorithmic}
   
\footnotesize

    \STATE Initialize $x, L$  \COMMENT{Start with $x$ as some rough guess for the parameters and $L$ large. }
\FOR{$t \gets 0$ to $T$}
\STATE $B = B_0/\text{tr}(LL^{\top})^{\kappa/2}$
\STATE Choose $B$ random values for the random vector $v$
\STATE SAMPLE $y_i = \hat{f}(Lx + v_i)$ for each random vector (can be parrallelized)
\STATE Using $y_i$, compute the estimates of $\frac{\partial h(L,x)}{\partial L} \approx \hat{h}_L$ and $\frac{\partial h(L,x)}{\partial x} \approx \hat{h}_x$ (equations \eqref{eq: est_L}, \eqref{eq: est_x})
\STATE Compute $\Delta L = \alpha_L L L^{\top}  \hat{h}_L$, $\Delta x = \alpha_x L L^{\top} \hat{h}_x$ (equations \eqref{eq: SDE})
\STATE Set $\hat{L} \leftarrow L + \Delta t \Delta L$
\STATE Set $\hat{\Delta t} \leftarrow \Delta t \left(\frac{|\hat{L}|}{|L|} \right)^{1/2}$ (for numerical stability)
\STATE Set $L \leftarrow L + \hat{\Delta t} \Delta L$, $x  \leftarrow x + \hat{\Delta t} \Delta x$ (update window)
\STATE Constrain $L \leftarrow \text{CLAMP}(L)$ (for numerical stability, see equation \eqref{eq: clamp} for definition of $\text{CLAMP}$)
\ENDFOR
\STATE \textbf{return $x$} \COMMENT{Return the putative best parameters}


\normalsize
\end{algorithmic}
\caption{\label{alg: SDE_tuner}}
\end{algorithm}

Although the algorithm is directly based on the numerical integration of eq. \eqref{eq: SDE}, there are several additional steps that we add to improve numerical stability. 
These steps are mainly to ensure that the window size does not grow or shrink too quickly during the optimal region search. We introduce the notation $|L| = \text{tr}(LL^{\top})^{1/2}$ to denote the norm of $L$ which also represents the size of the window.
\\
\\

The first step is to use something like a two-step integration scheme when updating the window matrix $L$. First, we compute $\hat{L} = L + \Delta t \Delta L$ (where $\Delta L$ is the estimate of $\partial L / \partial t$) using the original time step. Then we compute a new time step:
$\hat{\Delta t} \leftarrow \Delta t \frac{|\hat{L}|}{|L|}$
which we then use to update $L$ and $x$. The purpose of is to avoid the term $\Delta t \Delta L$ to become too large and negative which, in turn, will make $L$ shrink too quickly in a single time step.
\\
\\
Second, we limit the magnitude of $L$ so that it cannot shrink or grow too quickly. To do this, the following clamp function is applied to $L$ after each time-step:
\begin{equation}\label{eq: clamp}
    \text{CLAMP}(L) = 
        \begin{cases} 
            L \frac{w_{max} D^{1/2}}{|L|} & |L|/D^{1/2} > w_{max} \\
            L & w_{min} \leq |L|/D^{1/2} \leq w_{max} \\
            L \frac{w_{min} D^{1/2}}{|L|} & |L|/D^{1/2} < w_{min} 
        \end{cases}
\end{equation}
where $w_{min}$ and $w_{max}$ are the minimum and maximum window sizes respectively. For the numerical results in this work, we always use $w_{max} = 2$, and, except for the benchmark results on the Rosenbrock function (section \ref{sec: rosenbrock}), we use $w_{min} = 0$.

\clearpage

\subsection{Additional Numerical Results on artificial problems}\label{sec: additional_numres}

\begin{figure}
    \centering
    \includegraphics[width=0.45\textwidth]{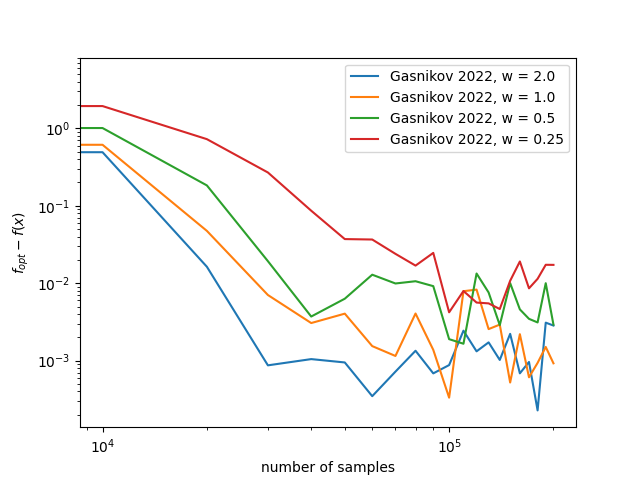}
    \includegraphics[width=0.45\textwidth]{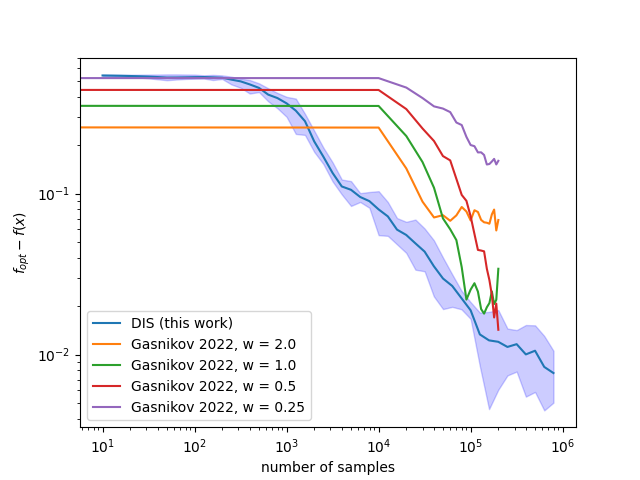}
    
    \caption{Left: Distance from true optimum plotted against the number of oracle calls for the algorithm analyzed in \cite{Gasnikov2022} for different window sizes/sampling radii. The fitness function used is a 5-dimensional Gaussian function (which is symmetric). Right: $f_{opt} - f(x)$ is plotted for both \cite{Gasnikov2022} and algorithm 3.3. The fitness function is a piece-wise parabolic function designed to be skewed and not have reflectional symmetry. }
    \label{fig: dynamic_size_motivation}
\end{figure}

\subsection{Window Size Tradeoff}\label{sec: tradeoff}

This trade-off is demonstrated numerically in Fig. \ref{fig: dynamic_size_motivation} of appendix section \ref{sec: additional_numres} in which two artificial fitness functions are optimized using the ball smoothing algorithm from \cite{Gasnikov2022}. In the leftmost plot, we use a symmetric fitness function $f(x) = 1 - \frac{1}{D}\sum_i x_i^2$ with Gaussian noise added at each measurement. Because this function is symmetric around its optimum, a smoothed version of the function will have the same optimum, and ball smoothing can use an arbitrarily large sampling radius. Thus, the largest sampling radius ($w = 2$) is most quickly able to obtain a nearly optimal configuration.
\\
\\
On the other hand, in the right plot of Fig. \ref{fig: dynamic_size_motivation} we use a different, asymmetric, fitness function $f(x) = 1 - \frac{1}{D}\sum_i (1 + 0.9\text{sign}(x_i))x_i^2$. This fitness function is designed to be highly asymmetric and also not locally quadratic so it is asymmetric on small scales as well. 
We test both the ball smoothing algorithm from \cite{Gasnikov2022} as well as the DIS from section \ref{sec: dynamic_ball_smoothing}. Because of the skewed fitness function, we can see the trade-off between large and small window sizes for Gasnikov's algorithm\cite{Gasnikov2022}. Although a large window size such as $w=2.0$ converges faster, it converges to a sub-optimal point and the fitness does not improve after $10^4$ samples. This motivates the use of a dynamic window size. In fact, we can see in Fig. \ref{fig: dynamic_size_motivation} that the DIS is always able to achieve a more optimal solution in a smaller number of samples on this fitness function. We can also see in this figure that because of the linear nature of the trace in the log-log plot that the asymptotic convergence of the error should be something like $O(n_s^{-c})$ for some $c$. In fact, in appendix \ref{sec: asymptotic} we show numerically that the error of DAS asymptotically scales as $O(D/n_s^{1/2})$ which is known to be optimal for type of noisy optimization \cite{Larson2019}.

\subsection{Heterogeneous curvature} \label{sec: window_shape}

\begin{figure}[h]
    \centering
    \includegraphics[width=0.45\textwidth]{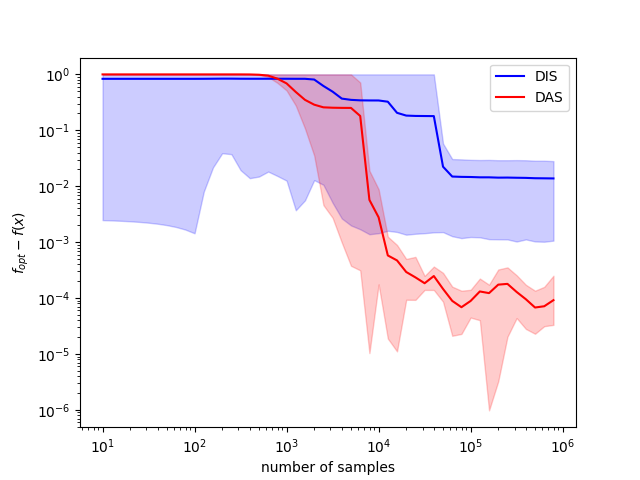}
    \includegraphics[width=0.45\textwidth]{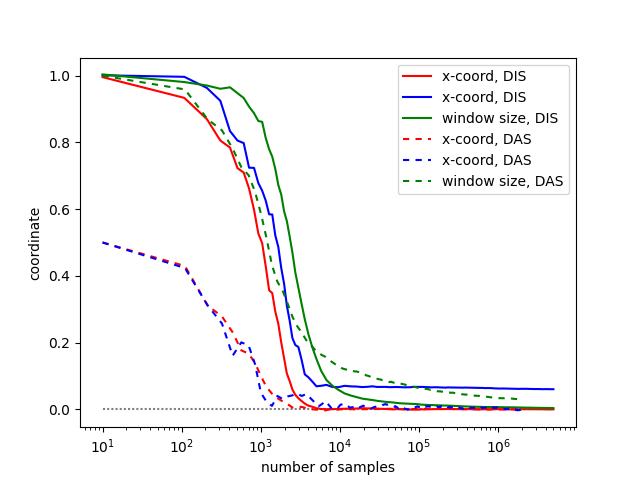}

    \caption{Left: Error of DIS with respect to number of samples. Right: Single trajectory of DAS demonstrating the reason for the failure of the circular window.}
    \label{fig: dynamic_shape_motivation}
\end{figure}

When tuning parameters, it is often the case that some parameters will have much higher sensitivity than others. This is true for the CO solvers tuned in this paper (see sections \ref{sec: CO} and \ref{sec: ising}) and has been commonly observed when tuning the weights of neural networks in machine learning\cite{Bollapragada2019adaptive,Yao2021adahessian,Liu2023sophia}. This phenomenon of heterogeneous curvature is the motivation behind changing the sampling window shape as described in section \ref{sec: dynamic_ellipsoid_smoothing}.
\\
\\
In Fig. \ref{fig: dynamic_shape_motivation}, we show numerical results on the 2-dimensional fitness function $f(x,y) = e^{-100x^2 - y^2}$. When a circular sampling window is used, the window size shrinks too quickly because of the sharp curvature in the $x$ direction. Because the window simultaneously shrinks in the $y$ direction, the algorithm struggles to accurately tune the less sensitive $y$ coordinate. However, if an elliptical sampling window is used, the algorithm correctly matches the shape of the objective function and treats each dimension appropriately.

\begin{figure}[h]
    \centering
    \includegraphics[width=0.4\textwidth]{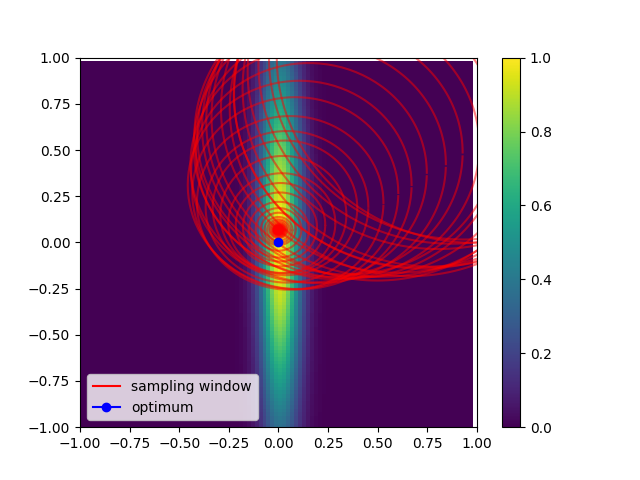}
    \includegraphics[width=0.4\textwidth]{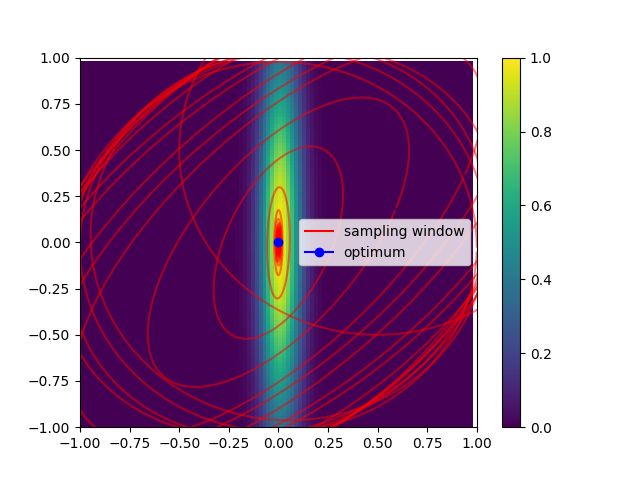}

    \caption{Left: Red circles show sampling window (representing one standard deviation of the Gaussian kernel) at different points in the algorithm on top of the heatmap of the fitness function (formula shown in text). Right: Red ellipses show sampling window of DAS at different points during the algorithm's trajectory. Data for both plots are the same as the right plot of figure \ref{fig: dynamic_shape_motivation}.   }
    \label{fig: window_trajectories}
\end{figure}

\subsection{Asymptotic Convergence}\label{sec: asymptotic}

In this section, we will numerically study the asymptotic convergence rate of DAS. Let $\epsilon(n_s)$ be the residual error in finding the true optimum of $f$ with $\epsilon(n_s) = f_{opt} - f(x)$ obtained after $n_s$ samples. Then at best, $\epsilon(n_s) = O(n_s^{-1/2})$\cite{Jamieson2012}. This is roughly because, due to the central limit theorem, we will need at least $\frac{\sigma^2}{\epsilon^2}$ samples to resolve $f$ to within $\epsilon$ at a single point. Many algorithms have been shown theoretically to achieve similar convergence rates on certain classes of problems (see table 8.1 of \cite{Larson2019}). 

\begin{figure*}[h]
    \centering
    \includegraphics[width=0.4\textwidth]{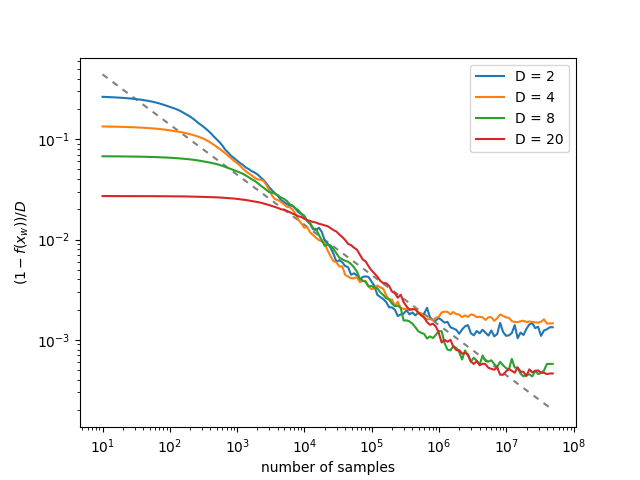}
    \includegraphics[width=0.4\textwidth]{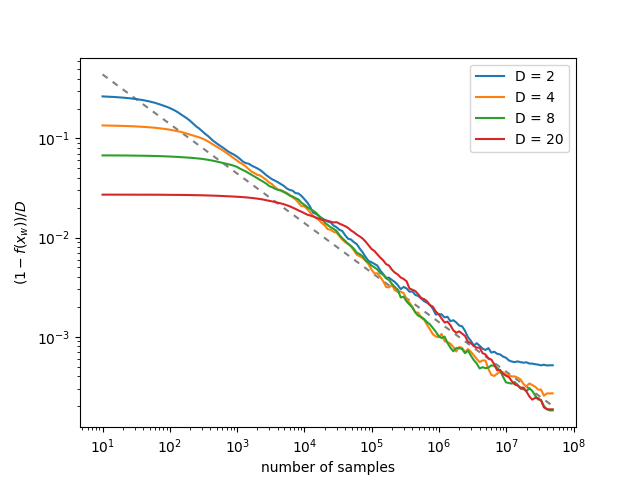}
    \caption{Normalized error,  $\epsilon/D$, of algorithm \ref{alg: SDE_tuner} as a function of the number of samples ($n_s$) for different dimensions. The dotted gray line represents the function $c/n_s^{1/2}$ with $c = 1.4$. Left: algorithm parameter $\kappa = 0.5$ , right $\kappa = 1.0$. Error is averaged over 5 runs for each trace. Fitness function is defined as $f(x) = 1 - \frac{1}{D}\sum_i (1 + 0.9\text{sign}(x_i))x_i^2$. }
    \label{fig: convergence}
\end{figure*}

In Fig. \ref{fig: convergence}, we show the error of algorithm \ref{alg: SDE_tuner} with two different values of $\kappa$. The fitness function used for these results is $f(x) = 1 - \frac{1}{D}\sum_i (1 + 0.9\text{sign}(x_i))x_i^2$ and the measurement noise is a zero mean Gaussian with variance $0.01$. Although $\kappa = 0.5$ has been used for many of the previous numerical results in \ref{sec: num_results} $\kappa = 1.0$ appears to show the correct asymptotic convergence in the large $n_s$ limit. When $\kappa = 1.0$ the algorithm appears to converge according to $\epsilon = O\left(\frac{D}{n_s^{1/2}}\right)$.

\clearpage

\subsection{Additional Results on Modified Rosenbrock Function}\label{sec: rosen_tables}
Tables \ref{tab: rosenD2} and \ref{tab: rosenD8} show results from figure \ref{fig: rosenbrock_bench} in tabular form. Bolded numbers represent the best fitness of each column. BOHB performs better in many cases, however, in higher dimensions, DAS will always outperform BOHB with enough samples.

\begin{table}[h]
    \centering
    \begin{small}
    
    \begin{tabular} {|p{0.08\columnwidth}||p{0.08\columnwidth} p{0.08\columnwidth} p{0.08\columnwidth}||p{0.08\columnwidth} p{0.08\columnwidth} p{0.08\columnwidth}||p{0.08\columnwidth} p{0.08\columnwidth} p{0.08\columnwidth}|}
        \toprule
        & $n_s = 10^3$ mean & $n_s = 10^3$ worst & $n_s = 10^3$ best& $n_s = 10^4$ mean & $n_s = 10^4$ worst & $n_s = 10^4$ best& $n_s = 10^5$ mean & $n_s = 10^5$ worst & $n_s = 10^5$ best \\ \hline
\underline{DAS} (this work)  & 0.734 & 0.549 & 0.852 & 0.925 & 0.861 & 0.981 & 0.993 & 0.982 & 0.997 \\ 
\underline{Gasnikov}, w=0.25  & 0.141 & 0.000 & 0.551 & 0.451 & 0.000 & 0.779 & 0.588 & 0.000 & 0.776 \\ 
\underline{SPSA} (Spall 1998)  & 0.270 & 0.001 & 0.652 & 0.800 & 0.662 & 0.966 & 0.990 & \textbf{0.986} & 0.998 \\ 
\underline{BOHB} (Falkner 2018)  & \textbf{0.902} & \textbf{0.775} & \textbf{0.996} & \textbf{0.963} & \textbf{0.868} & \textbf{0.999} & \textbf{0.994} & 0.980 & \textbf{0.999} \\  \bottomrule
    \end{tabular} 
    
    \caption{Table shows best, worst, and mean fitness achieved by four algorithms for different values of $n_s$. The artificial problem is the modified Rosenbrock function in 2 dimensions with $\beta = 0.5$. }
    \label{tab: rosenD2}
    \end{small}
\end{table}

\begin{table}[h]
    \centering
    \begin{small}
    
    \begin{tabular} {|p{0.08\columnwidth}||p{0.08\columnwidth} p{0.08\columnwidth} p{0.08\columnwidth}||p{0.08\columnwidth} p{0.08\columnwidth} p{0.08\columnwidth}||p{0.08\columnwidth} p{0.08\columnwidth} p{0.08\columnwidth}|}
        \hline 
      & $n_s = 10^4$ mean & $n_s = 10^4$ worst & $n_s = 10^4$ best& $n_s = 10^5$ mean & $n_s = 10^5$ worst & $n_s = 10^5$ best& $n_s = 10^6$ mean & $n_s = 10^6$ worst & $n_s = 10^6$ best \\ \hline
      
\underline{DAS} (this work)  & 0.000 & 0.000 & 0.000 & 0.000 & 0.000 & 0.000 & \textbf{0.192} & 0.000 & \textbf{0.962} \\ \hline
\underline{Gasnikov}, w=0.25  & 0.000 & 0.000 & 0.000 & \textbf{0.083} & 0.000 & \textbf{0.415} & 0.044 & 0.000 & 0.222 \\ \hline
\underline{SPSA} (Spall 1998)  & 0.000 & 0.000 & 0.000 & 0.000 & 0.000 & 0.000 & 0.000 & 0.000 & 0.000 \\ \hline
\underline{BOHB} (Falkner 2018)  & \textbf{0.031} & 0.000 & \textbf{0.085} & 0.039 & \textbf{0.002} & 0.083 & 0.039 & \textbf{0.002} & 0.083 \\ 
\hline
    
    \end{tabular}
    
    \caption{Table shows best, worst, and mean fitness achieved by four algorithms for different values of $n_s$. The artificial problem is the modified Rosenbrock function in 8 dimensions with $\beta = 0.2$. }
    \label{tab: rosenD8}
    
    \end{small}
\end{table}

\clearpage

\section{Details on Combinatorial Optimization Solvers}\label{sec: co_details}

\subsection{Additional Discussion on Combinatorial Optimization}\label{sec: co_disc}

A common problem with all heuristic CO solvers, especially the differential solvers, is that there typically are many parameters that need to be optimized to ensure good performance \cite{Leleu2019, Reifenstein2021, Reifenstein2023, Goto2021, Kalinin2023}. Additionally, the optimal parameters can be very different depending on the exact type of problem that is being solved (e.g. for some application-specific problem) and often vary from instance to instance as well. The problem of parameter make it hard to fairly compare different heuristics. The main motivation behind the tuning algorithm developed in this work is to be able to quickly and accurately find good parameters for these new differential solvers. This will allow us to further the development and benchmarking of them. Improving the parameter selection techniques for differential solvers will possibly allow them to be more competitive against classical heuristics in situations where they currently struggle.

The problem of tuning parameters for combinatorial optimization is not new. For example, many previous works have developed methods for tuning heuristic SAT solvers  \cite{Hutter2007ParamILS, Hutter2011SATTune, Hoos2021, Fuchs2023SatTune}. These works focused on tuning both discrete and continuous parameters. Additionally, these works tend to use very basic methods for optimizing the continuous parameters. In our work, we focus on continuous parameters only and base our optimizer on previous results in derivative-free optimization. 
\\
\\
More recently, due to the growing interest in Ising machines, several authors have studied methods for tuning Ising machines \cite{Parizy2023, Bernal2021}. Unlike our methods, these methods are more closely related to bayesian optimization. \cite{Parizy2023}  uses a Parzan tree estimator similar to that of BOHB \cite{BOHB} while \cite{Bernal2021} is directly based on Hyperopt \cite{Hyperopt}. 
\\
\\
Additionally, many authors have attempted to use machine learning directly to help solve optimization problems. For example, graph neural networks\cite{Schuetz2022, Selsam2019} and deep neural networks\cite{Taassob2023} have been used to learn the solutions of combinatorial optimization problems. Although these works may seem not so related at first glance, the tuning of parameters for a differential solver can be interpreted as training a recurrent graph neural network with a very small dimensional parameter space. This work touches on a deep connection between graph neural networks, reinforcement learning, differential solvers, and Ising machines.

\clearpage

\subsection{QUBO/Ising CAC}\label{sec: ising_details}

Coherent Ising machines (CIMs) are a proposed method for solving the Ising problem in which the Ising spins are represented with analog amplitudes. 
Although originally proposed as an optical analog computer \cite{Wang2013}, the CIM can be simulated by numerical integration of an ODE. These ODEs typically have several parameters that need to be tuned precisely for the device to have good performance on a particular type of Ising problem.
\\
\\
The current state-of-the-art CIM algorithm is defined by the following ODE \cite{Leleu2019} which we will refer to as CIM-CAC (CIM with chaotic amplitude control).
\begin{equation}\label{eq: CAC1}
    \frac{dx_i}{dt} = x_i(p - 1 - x_i^2) - e_i \sum_{j} J_{ij} x_j
\end{equation}
\begin{equation}\label{eq: CAC2}
    \frac{de_i}{dt} = \beta e_i \left(1 - x_i^2 \right)
\end{equation}
For an Ising problem of size $N$ and coupling matrix $J_{ij}$, the algorithm evolves both $N$ dimensional vectors $x$ and $e$ over time using an Euler numerical integration step. 
The variables are initialized randomly, and the sign of $x$ represents the possible solution of the corresponding Ising problem. Typically many trajectories are used to find a good solution. The parameters $\beta$ and $p$ as well as the numerical integration step $\text{dt}$ are important to tune precisely for the algorithm to be effective.

\subsection{SAT-CAC}

The coherent SAT solver is designed to find a variable assignment that satisfies the problem, or, if it is unsatisfiable, find an assignment that satisfies the maximum number of clauses (MAX-SAT). The SAT problem is specified by a sparse matrix $C_{ij}$ as follows:
\begin{equation}
C_{ij} =\left\{
\begin{array}{ll}
      1 &  i\text{th variable is included un-negated in }j\text{th clause} \\
      -1 & i\text{th variable is included negated in }j\text{th clause} \\
      0 & i\text{th variable is not included in }j\text{th clause} \\
\end{array} 
\right.
\end{equation}
Then we also define the set $I_j$ for each clause as $I_j = \left\{ i \mid C_{ij} \neq 0 \right\}$
The boolean variables are represented by soft spins $x_i \in \mathbb{R}$ where $x_i > 0$ represents True and $x_i < 0$ represents False. Then we define the following quantities:
\begin{equation}
K_j = \prod_{i \in I_j} \frac{1 - C_{ij}x_i}{2} \quad K_{ij} = \frac{-C_{ij}}{2}\prod_{k \in I_j, k \neq i} \frac{1 - C_{kj}x_k}{2} 
\end{equation}
Then, the coherent SAT solver equations can be written as:
\begin{equation}
\frac{dx_i}{dt} = x_i (p - 1 - x_i^2) - e_i \sum_j K_{ij}
\end{equation}
\begin{equation}
\frac{de_i}{dt} = \beta e_i (1 - x_i^2)
\end{equation}
These equations are nearly identical to the CIM equations \eqref{eq: CAC1}\eqref{eq: CAC2} and thus have the same system parameters. 
Additionally, the parameter $p$ is typically set to change throughout the trajectory, in total there are four relevant parameters, $\text{dt}$, $p_{init}$, $p_{end}$, and $\beta$.

\clearpage

\subsection{Tuning Trajectories for SAT Solver}\label{sec: tune_traj}
\begin{figure*}[h]
   
    \centering
    \includegraphics[width = 0.3 \textwidth]{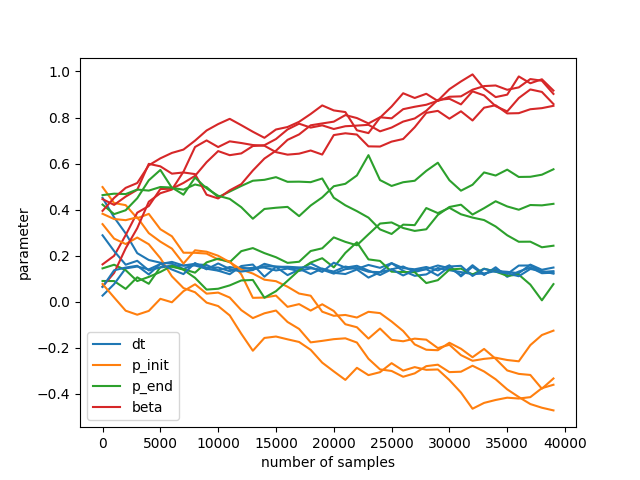}
    \includegraphics[width = 0.3 \textwidth]{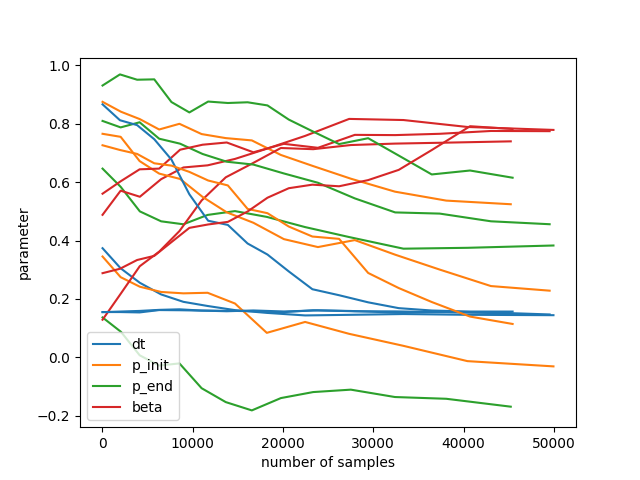}
    \includegraphics[width = 0.3 \textwidth]{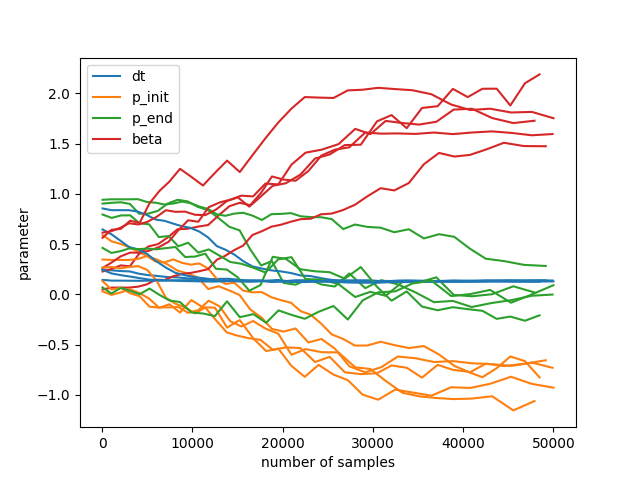}
    
    \caption{Tuning trajectories over 4-5 different initial conditions for Gasnikov et al 2022 (left), DIS (middle), and DAS (right). }
    \label{fig: SAT_tune_trajectories}
\end{figure*}

In Fig. \ref{fig: SAT_tune_trajectories}, we show the trajectories of each parameter during the tuning process for three algorithms. The first algorithm (following \cite{Gasnikov2022}) uses a fixed sampling window with a fixed size. The second one uses a sampling window with a dynamic size but fixed shape and the third uses a dynamic size and shape as described in this work.
Both the first and third algorithms converge on roughly the same parameters regardless of the initial condition, while the middle does not. This is because (as described earlier) different sensitivities to different parameters can cause the sampling window to prematurely shrink. In this case, the $\text{dt}$ parameter is much more sensitive, so the algorithm will tune $\text{dt}$ while neglecting the others. With a dynamic window shape, this is not a problem and we can see that all parameters appear to be tuned accurately. On the other hand, even though the Gasnikov algorithm is consistent with the parameters it chooses, these parameters are sub-optimal (see figure \ref{fig: SAT_comparison}
) because it is operating on a smoothed version of the objective function. 

\clearpage

\subsection{Additional Results on QUBO/Ising} \label{sec: ising}

\begin{figure}[b]
   
    \centering
    \includegraphics[width = 0.4 \textwidth]{figures/Ising_comparison_150N.png}
    s\includegraphics[width = 0.4 \textwidth]{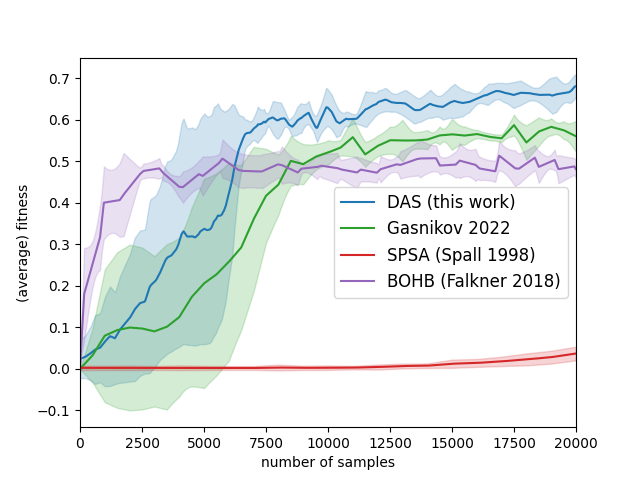}
    
    \caption{Average fitness (soft success rate) obtained by different tuning methods on random SK model instance with, $N = 150, \beta_E = 0.01$ (left) and $N = 300, \beta_E = 0.005$ (right). The fitness function is defined by equation \eqref{eq: soft_ps}. Averages are over 5 realizations of the tuning dynamics starting at different randomized positions. The shaded region represents one standard deviation of the data. To evaluate the fitness for each parameter configuration, 20 random SK instances are generated, and 50 trajectories are evaluated for each. The left plot of this figure is identical to the right plot of \ref{fig: SAT_comparison} of the main text but is included here as well for completeness.}
    \label{fig: Ising_comprison}
    
\end{figure}
\begin{figure}
   
    \centering
    \includegraphics[width = 0.4 \textwidth]{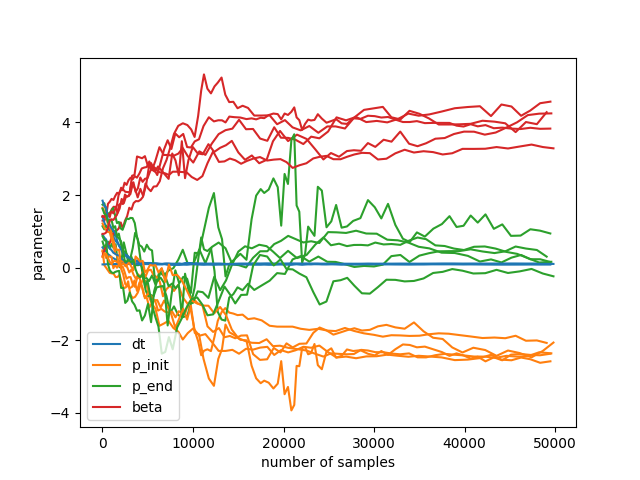}
    \includegraphics[width = 0.4 \textwidth]{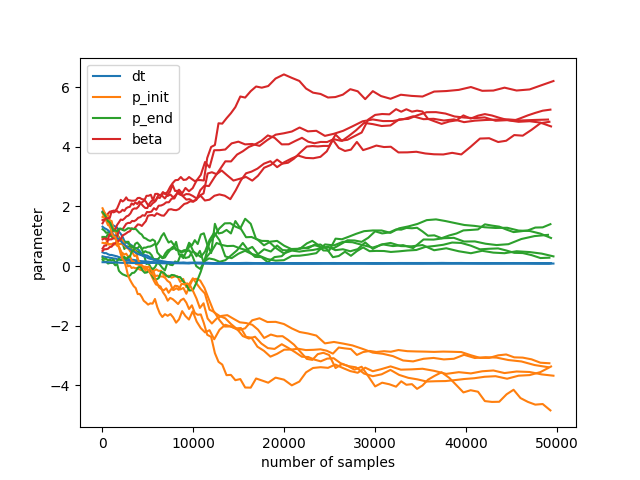}
    
    \caption{Parameters obtained by tuning CIM-CAC with DAS on random SK model instance with, $N = 150, \beta_E = 0.01$ (left) and $N = 300, \beta_E = 0.005$ (right). The fitness function is defined by equation \eqref{eq: soft_ps}. 5 realizations of the tuning dynamics starting at different randomized positions are shown.  }
    \label{fig: Ising_tune_traj}
    
\end{figure}

In this section, we consider tuning of the coherent Ising machine, which is described in detail in section \ref{sec: ising_details}. A common way that Ising machines are benchmarked has been by randomly choosing values for the weight matrix. This ensemble of problem instances is known as the Sherrington-Kirkpatrick (SK) model in spin glass physics \cite{SK}. The SK model has been very well studied from the perspective of statistical physics and spin glass theory and this is one of the reasons it is commonly used as a benchmark for Ising machines. For example, the average ground state energy of an SK problem instance can be accurately approximated \cite{Boettcher2010, Parisi1980}. For an SK problem instance with entries chosen from a zero mean Gaussian the ground state will on average be:
\begin{equation}
    E_{avg} \approx N^{3/2}(-0.761 + 0.7 N^{-2/3})
\end{equation}
ignoring some higher order finite size corrections \cite{Boettcher2010}. This is useful because when we randomly generate these instances we do not know their ground state or ground state energy. So, unlike with the SAT problem, we cannot use a success rate as our fitness function since we do not know when the algorithm is successful or not. However, this formula gives us a rough estimate of the ground state energy which we can use to form a different type of fitness function we will call a "soft success rate" as follows:
\begin{equation}\label{eq: soft_ps}
    f(x) = \mathbb{E}\left(e^{-\beta_E (\hat{E} - E_{thresh})} \right)
\end{equation}
This formula, which is inspired by the Boltzmann factor for statistical mechanics, depends on an inverse temperature parameter $\beta_E$ (not to be confused with the $\beta$ in equation \eqref{eq: CAC2}). $\hat{E}$ is the Ising energy returned by a single trajectory for CIM-CAC and $E_{thresh}$ is set to $N^{3/2}(-0.761 + 0.7 N^{-2/3})$. The expected value is over both initial conditions for CIM-CAC and the ensemble of SK instances. If $\beta_E$ is small, then this cost function is similar to the average energy returned by the solver, while if $\beta_E$ is large then the fitness is closer to a success rate. This success rate is not the success rate of finding the ground state but rather the success rate of finding energy under a threshold energy for a given instance. If we are interested in finding the ground state, then we will have to assume that the optimal parameters for this cost function are also useful for finding the ground state. On the other hand, if we are just interested in finding low energy states efficiently then this formula makes sense and we can choose $\beta_E$ appropriately.
\\
\\
A more in-depth study is needed to understand the properties and usefulness of this type of fitness function. In this work, we will use this cost function as an example of how our proposed algorithm can be used for tuning an Ising machine. In figure \ref{fig: Ising_comprison} we show the fitness of four tuning algorithms as a function of the number of samples. Again, BOHB \cite{BOHB} and \cite{Gasnikov2022} with sampling window set to $w=2$ are both effective at tuning the solver. However, similar to the case of SAT (figure \ref{fig: SAT_comparison}) DAS is consistently able to achieve better fitness given enough samples. This is mainly because of its ability to adjust to the different sensitivities of the different parameters. Additionally, we see in figure \ref{fig: Ising_tune_traj} that DAS is consistent at finding the same parameters at both problem sizes. Similar to the SAT case (figure \ref{fig: SAT_tune_trajectories} the optimal parameters appear to have very extremal values, however DAS is still able to find them accurately.
\\
\\

\clearpage

\section{Comparison with Previous Methods}\label{sec: discussion}

\subsection{Comparison to other sampling window approaches}

The most common approach to derivative-free optimization is to replace the true objective function with a "smoothed" version defined as
$$f_{\gamma}(x) = \gamma \int \kappa( (x-u)/ \gamma) f(u)du$$
where $\kappa$ is some convolution kernel (typically a Gaussian or a step function) \cite{Nesterov2017,Gasnikov2022, Gasnikov2022R}. By sampling $f$ in a region of size $\gamma$ around $x$ we can approximate the gradient of $f_{\gamma}$. When $f$ is non-differentiable or when it cannot be measured without a small amount of noise this technique is powerful because it is able to smooth over these difficulties so a traditional gradient descent method can be used. 

stochastic approximation (SA) methods such as \cite{Spall1998} approximate the gradient of $f$ with what is often called a stencil. That is we choose a random vector $\Delta$ from the set $\{-1,1\}^D$ and then approximate the gradient using the formula 
$$ g(x) = \Delta \frac{f(x_k + \Delta c_k) - f(x_k - \Delta c_k)}{2c_k}$$
Where $k$ is the iteration number, $x_k$ is the current position and $c_k$ is the current window size. 
The position is updated according to gradient descent (or ascent) as $x_{k+1} = x_k - a_k g(x_k)$. The sequence of numbers $c_k$ and $a_k$ are set to decay with iteration number based on some function depending on the exact implementation and parameters. In this work we use the SA method described in \cite{Spall1998}, which we will call SPSA, to represent algorithms in this class.

\subsection{Comparison to Bayesian optimization}

Algorithms based on Bayesian optimization typically make use of different budgets when evaluating the objective function. That is, less accurate less, computationally intensive evaluations are combined with more computationally intensive and accurate evaluations. In this work we will use BOHB to represent this class of tuners for benchmarking purposes since it is a commonly used algorithm in the field. For a more in-depth review of Bayesian (and other types of) hyperparameter tuners we refer to \cite{Bischi2023}.

\subsection{Comparison to Machine learning approaches using the Hessian \label{sec: hessian_ML_rev}}

Diagonal Hessian pre-conditioning has been used in neural network optimizer design\cite{Schaul2013} with recent iterations like AdaHessian\cite{Yao2021adahessian}. Several methods use the Gauss-Newton decomposition\cite{Ortega2000} to simplify the computing the pre-conditioners in second-order optimizers\cite{Botev2017,Gargiani2020}. Other recently regularization methods proposed that improve on ADAM includes decoupled weight decay (AdamW)\cite{oshchilov2017decoupled}, Lion\cite{Chen2023}, and stochastic variance reduction\cite{lan2018optimal}. In this paper, the focus is on tuning combinatorial heuristic but more extensive benchmarking with other methods for training neural network is the subject of future works.

\subsection{Differences and Similarities between Algorithms} \label{sc: diff_sim}

Although derivative-free optimization and hyper-parameter optimization are very well-developed fields, there are a number of distinct properties that our new method has which we believe make our new method unique and also optimal for the type of problems it is designed to solve. First of all, other than the Nedler-Mead-based methods, to the best of our knowledge out method is the first to use a sampling window that changes smoothly in both size and shape to fit the cost function. Having a dynamic window shape can be advantageous when some parameters have higher sensitivity than others and there are complex correlations between parameters. This is demonstrated numerically in sections \ref{sec: num_results}.
\\
\\
Another subtle important distinction between the different algorithms is how the samples of the objective functions are taken. Methods such as DS, TR, and Nedler-Mead which are originally designed for noiseless derivative-free optimization require evaluating the objective function with high accuracy at specific points. Additionally, Bayesian methods such as BOHB also require an accurate measurement of the objective function using a high computational budget. In our case, this means sampling the objective function many times at a specific point. On the other hand, ball smoothing and SA methods use many nosy samples at many different locations to approximate the gradient and do not need to directly compare the objective function at two locations using many samples. Our method, which is most closely related to ball ball smoothing \cite{Nesterov2017} falls into the second class of algorithms. When many noisy objective evaluations are available and we want to get very close to the optimum in a high dimensional parameter space we believe this second class of methods is better. This is because to resolve small differences in the objective function we will need to waste computation sampling at one location many times. On the other hand, if the samples are taken at different points we can simultaneously get information about both the objective function and its gradient.
\\
\\
Another reason that we base our algorithm on ball smoothing is that when it comes to our target applications we want to sample points over a continuous distribution of points such as a Gaussian. The reason for this is that for many parameter configurations CO solvers have a zero or close to zero success rate. For methods that use a discrete sampling window such as SA, DS or Nelder-Mead, it is likely that the initial sampling distribution completely misses the region of the parameter space where the objective function is nonzero. With our method, we can start with a large initial sampling window and sample points according to a continuous distribution until we by chance find a good objective evaluation in the region of interest.

Table \ref{tab: review} summarizes the similarities and differences between different derivative-free optimization methods that we will discuss.
    
    
    

\clearpage

\bibliography{main}

\begin{thebibliography}{80}
\providecommand{\natexlab}[1]{#1}
\providecommand{\url}[1]{\texttt{#1}}
\expandafter\ifx\csname urlstyle\endcsname\relax
  \providecommand{\doi}[1]{doi: #1}\else
  \providecommand{\doi}{doi: \begingroup \urlstyle{rm}\Url}\fi

\bibitem[Agarwal et~al.(2010)Agarwal, Dekel, and Xiao]{Agarwal2010}
Agarwal, A., Dekel, O., and Xiao, L.
\newblock Optimal algorithms for online convex optimization with multi-point bandit feedback.
\newblock In \emph{Annual Conference Computational Learning Theory}, 2010.
\newblock URL \url{https://api.semanticscholar.org/CorpusID:118314530}.

\bibitem[Akiba et~al.(2019)Akiba, Sano, Yanase, Ohta, and Koyama]{Optuna}
Akiba, T., Sano, S., Yanase, T., Ohta, T., and Koyama, M.
\newblock Optuna: A next-generation hyperparameter optimization framework, 2019.

\bibitem[Ans{\'o}tegui et~al.(2009)Ans{\'o}tegui, Sellmann, and Tierney]{GGA}
Ans{\'o}tegui, C., Sellmann, M., and Tierney, K.
\newblock A gender-based genetic algorithm for the automatic configuration of algorithms.
\newblock In \emph{International Conference on Principles and Practice of Constraint Programming}, pp.\  142--157. Springer, 2009.

\bibitem[Ans{\'o}tegui et~al.(2015)Ans{\'o}tegui, Malitsky, Samulowitz, Sellmann, Tierney, et~al.]{GGAp}
Ans{\'o}tegui, C., Malitsky, Y., Samulowitz, H., Sellmann, M., Tierney, K., et~al.
\newblock Model-based genetic algorithms for algorithm configuration.
\newblock In \emph{IJCAI}, pp.\  733--739, 2015.

\bibitem[Barton \& Ivey(1996)Barton and Ivey]{Barton1996}
Barton, R.~R. and Ivey, J.~S.
\newblock Nelder-mead simplex modifications for simulation optimization.
\newblock \emph{Management Science}, 42\penalty0 (7):\penalty0 954--973, 1996.
\newblock \doi{10.1287/mnsc.42.7.954}.
\newblock URL \url{https://doi.org/10.1287/mnsc.42.7.954}.

\bibitem[Berahas et~al.(2022)Berahas, Cao, Choromanski, and Scheinberg]{Berahas2022}
Berahas, A.~S., Cao, L., Choromanski, K., and Scheinberg, K.
\newblock A theoretical and empirical comparison of gradient approximations in derivative-free optimization.
\newblock \emph{Foundations of Computational Mathematics}, 22\penalty0 (2):\penalty0 507--560, 2022.

\bibitem[Bergstra et~al.(2013)Bergstra, Yamins, and Cox]{Hyperopt}
Bergstra, J., Yamins, D., and Cox, D.
\newblock Making a science of model search: Hyperparameter optimization in hundreds of dimensions for vision architectures.
\newblock In Dasgupta, S. and McAllester, D. (eds.), \emph{Proceedings of the 30th International Conference on Machine Learning}, volume~28 of \emph{Proceedings of Machine Learning Research}, pp.\  115--123, Atlanta, Georgia, USA, 17--19 Jun 2013. PMLR.
\newblock URL \url{https://proceedings.mlr.press/v28/bergstra13.html}.

\bibitem[Bernal(2021)]{Bernal2021}
Bernal, D.
\newblock Window sticker - stochastic benchmark.
\newblock \url{https://github.com/usra-riacs/stochastic-benchmark}, 2021.

\bibitem[Birattari et~al.(2010)Birattari, Yuan, Balaprakash, and St{\"u}tzle]{f-race}
Birattari, M., Yuan, Z., Balaprakash, P., and St{\"u}tzle, T.
\newblock F-race and iterated f-race: An overview.
\newblock \emph{Experimental methods for the analysis of optimization algorithms}, pp.\  311--336, 2010.

\bibitem[Bischl et~al.(2023)Bischl, Binder, Lang, Pielok, Richter, Coors, Thomas, Ullmann, Becker, Boulesteix, Deng, and Lindauer]{Bischi2023}
Bischl, B., Binder, M., Lang, M., Pielok, T., Richter, J., Coors, S., Thomas, J., Ullmann, T., Becker, M., Boulesteix, A.-L., Deng, D., and Lindauer, M.
\newblock Hyperparameter optimization: Foundations, algorithms, best practices, and open challenges.
\newblock \emph{WIREs Data Mining and Knowledge Discovery}, 13\penalty0 (2):\penalty0 e1484, 2023.
\newblock \doi{https://doi.org/10.1002/widm.1484}.
\newblock URL \url{https://wires.onlinelibrary.wiley.com/doi/abs/10.1002/widm.1484}.

\bibitem[Boettcher(2010)]{Boettcher2010}
Boettcher, S.
\newblock Simulations of ground state fluctuations in mean-field ising spin glasses.
\newblock \emph{Journal of Statistical Mechanics: Theory and Experiment}, 2010\penalty0 (07):\penalty0 P07002, jul 2010.
\newblock \doi{10.1088/1742-5468/2010/07/P07002}.
\newblock URL \url{https://dx.doi.org/10.1088/1742-5468/2010/07/P07002}.

\bibitem[Bogolubsky et~al.(2016)Bogolubsky, Dvurechenskii, Gasnikov, Gusev, Nesterov, Raigorodskii, Tikhonov, and Zhukovskii]{Bogolubsky2016}
Bogolubsky, L., Dvurechenskii, P., Gasnikov, A., Gusev, G., Nesterov, Y., Raigorodskii, A.~M., Tikhonov, A., and Zhukovskii, M.
\newblock Learning supervised pagerank with gradient-based and gradient-free optimization methods.
\newblock \emph{Advances in neural information processing systems}, 29, 2016.

\bibitem[Bollapragada \& Wild(2019{\natexlab{a}})Bollapragada and Wild]{Bollapragada2019}
Bollapragada, R. and Wild, S.~M.
\newblock Adaptive sampling quasi-newton methods for derivative-free stochastic optimization.
\newblock \emph{arXiv preprint arXiv:1910.13516}, 2019{\natexlab{a}}.

\bibitem[Bollapragada \& Wild(2019{\natexlab{b}})Bollapragada and Wild]{Bollapragada2019adaptive}
Bollapragada, R. and Wild, S.~M.
\newblock Adaptive sampling quasi-newton methods for derivative-free stochastic optimization.
\newblock \emph{arXiv preprint arXiv:1910.13516}, 2019{\natexlab{b}}.

\bibitem[Botev et~al.(2017)Botev, Ritter, and Barber]{Botev2017}
Botev, A., Ritter, H., and Barber, D.
\newblock Practical gauss-newton optimisation for deep learning.
\newblock In \emph{International Conference on Machine Learning}, pp.\  557--565. PMLR, 2017.

\bibitem[Boyd \& Vandenberghe(2004)Boyd and Vandenberghe]{Boyd2004convex}
Boyd, S.~P. and Vandenberghe, L.
\newblock \emph{Convex optimization}.
\newblock Cambridge university press, 2004.

\bibitem[Chen et~al.(2023)Chen, Liang, Huang, Real, Wang, Liu, Pham, Dong, Luong, Hsieh, et~al.]{Chen2023}
Chen, X., Liang, C., Huang, D., Real, E., Wang, K., Liu, Y., Pham, H., Dong, X., Luong, T., Hsieh, C.-J., et~al.
\newblock Symbolic discovery of optimization algorithms.
\newblock \emph{arXiv preprint arXiv:2302.06675}, 2023.

\bibitem[Choromanski et~al.(2018)Choromanski, Iscen, Sindhwani, Tan, and Coumans]{Choromanski2018}
Choromanski, K., Iscen, A., Sindhwani, V., Tan, J., and Coumans, E.
\newblock Optimizing simulations with noise-tolerant structured exploration.
\newblock In \emph{2018 IEEE International Conference on Robotics and Automation (ICRA)}, pp.\  2970--2977. IEEE, 2018.

\bibitem[Conn et~al.(2000)Conn, Gould, and Toint]{Conn2000}
Conn, A., Gould, N., and Toint, P.
\newblock \emph{Trust Region Methods}.
\newblock MOS-SIAM Series on Optimization. Society for Industrial and Applied Mathematics (SIAM, 3600 Market Street, Floor 6, Philadelphia, PA 19104), 2000.
\newblock ISBN 9780898719857.
\newblock URL \url{https://books.google.com/books?id=wfs-hsrd4WQC}.

\bibitem[Conn et~al.(2009)Conn, Scheinberg, and Vicente]{Conn2009introduction}
Conn, A.~R., Scheinberg, K., and Vicente, L.~N.
\newblock \emph{Introduction to derivative-free optimization}.
\newblock SIAM, 2009.

\bibitem[Demmel et~al.(2019)Demmel, Li, Lakhdar, and USDOE]{GPTune}
Demmel, J.~W., Li, S.~X., Lakhdar, M. W.~S., and USDOE.
\newblock Gptune v.1, 8 2019.
\newblock URL \url{https://www.osti.gov//servlets/purl/1569031}.

\bibitem[Deng \& Ferris(2006)Deng and Ferris]{Deng2006}
Deng, G. and Ferris, M.~C.
\newblock Adaptation of the uobyqa algorithm for noisy functions.
\newblock In \emph{Proceedings of the 38th Conference on Winter Simulation}, WSC '06, pp.\  312–319. Winter Simulation Conference, 2006.
\newblock ISBN 1424405017.

\bibitem[Domhan et~al.(2015)Domhan, Springenberg, and Hutter]{Domhan2015}
Domhan, T., Springenberg, J.~T., and Hutter, F.
\newblock Speeding up automatic hyperparameter optimization of deep neural networks by extrapolation of learning curves.
\newblock In \emph{Twenty-fourth international joint conference on artificial intelligence}, 2015.

\bibitem[Duchi et~al.(2015)Duchi, Jordan, Wainwright, and Wibisono]{Duchi2015}
Duchi, J.~C., Jordan, M.~I., Wainwright, M.~J., and Wibisono, A.
\newblock Optimal rates for zero-order convex optimization: The power of two function evaluations.
\newblock \emph{IEEE Transactions on Information Theory}, 61\penalty0 (5):\penalty0 2788--2806, 2015.

\bibitem[Ercsey-Ravasz \& Toroczkai(2011)Ercsey-Ravasz and Toroczkai]{ErcseyRavasz2011}
Ercsey-Ravasz, M. and Toroczkai, Z.
\newblock Optimization hardness as transient chaos in an analog approach to constraint satisfaction.
\newblock \emph{Nature Physics}, 7\penalty0 (12):\penalty0 966--970, 2011.
\newblock \doi{10.1038/nphys2105}.
\newblock URL \url{https://doi.org/10.1038/nphys2105}.

\bibitem[Falkner et~al.(2018)Falkner, Klein, and Hutter]{BOHB}
Falkner, S., Klein, A., and Hutter, F.
\newblock {BOHB}: Robust and efficient hyperparameter optimization at scale.
\newblock In Dy, J. and Krause, A. (eds.), \emph{Proceedings of the 35th International Conference on Machine Learning}, volume~80 of \emph{Proceedings of Machine Learning Research}, pp.\  1437--1446. PMLR, 10--15 Jul 2018.
\newblock URL \url{https://proceedings.mlr.press/v80/falkner18a.html}.

\bibitem[Frazier(2018)]{Frazier2018tutorial}
Frazier, P.~I.
\newblock A tutorial on bayesian optimization, 2018.

\bibitem[Fuchs(2023)]{Fuchs2023SatTune}
Fuchs, T.
\newblock Automated tuning and portfolio selection for sat solvers.
\newblock Master's thesis, Karlsruhe Institute of Technology, 2023.

\bibitem[Gargiani et~al.(2020)Gargiani, Zanelli, Diehl, and Hutter]{Gargiani2020}
Gargiani, M., Zanelli, A., Diehl, M., and Hutter, F.
\newblock On the promise of the stochastic generalized gauss-newton method for training dnns.
\newblock \emph{arXiv preprint arXiv:2006.02409}, 2020.

\bibitem[Gasnikov et~al.(2022{\natexlab{a}})Gasnikov, Dvinskikh, Dvurechensky, Gorbunov, Beznosikov, and Lobanov]{Gasnikov2022R}
Gasnikov, A., Dvinskikh, D., Dvurechensky, P., Gorbunov, E., Beznosikov, A., and Lobanov, A.
\newblock Randomized gradient-free methods in convex optimization, 2022{\natexlab{a}}.

\bibitem[Gasnikov et~al.(2022{\natexlab{b}})Gasnikov, Novitskii, Novitskii, Abdukhakimov, Kamzolov, Beznosikov, Takac, Dvurechensky, and Gu]{Gasnikov2022}
Gasnikov, A., Novitskii, A., Novitskii, V., Abdukhakimov, F., Kamzolov, D., Beznosikov, A., Takac, M., Dvurechensky, P., and Gu, B.
\newblock The power of first-order smooth optimization for black-box non-smooth problems.
\newblock In Chaudhuri, K., Jegelka, S., Song, L., Szepesvari, C., Niu, G., and Sabato, S. (eds.), \emph{Proceedings of the 39th International Conference on Machine Learning}, volume 162 of \emph{Proceedings of Machine Learning Research}, pp.\  7241--7265. PMLR, 17--23 Jul 2022{\natexlab{b}}.
\newblock URL \url{https://proceedings.mlr.press/v162/gasnikov22a.html}.

\bibitem[Goto et~al.(2019)Goto, Tatsumura, and Dixon]{Goto2019}
Goto, H., Tatsumura, K., and Dixon, A.~R.
\newblock Combinatorial optimization by simulating adiabatic bifurcations in nonlinear hamiltonian systems.
\newblock \emph{Science Advances}, 5\penalty0 (4):\penalty0 eaav2372, 2019.
\newblock \doi{10.1126/sciadv.aav2372}.
\newblock URL \url{https://www.science.org/doi/abs/10.1126/sciadv.aav2372}.

\bibitem[Goto et~al.(2021)Goto, Endo, Suzuki, Sakai, Kanao, Hamakawa, Hidaka, Yamasaki, and Tatsumura]{Goto2021}
Goto, H., Endo, K., Suzuki, M., Sakai, Y., Kanao, T., Hamakawa, Y., Hidaka, R., Yamasaki, M., and Tatsumura, K.
\newblock High-performance combinatorial optimization based on classical mechanics.
\newblock \emph{Science Advances}, 7\penalty0 (6):\penalty0 eabe7953, 2021.
\newblock \doi{10.1126/sciadv.abe7953}.
\newblock URL \url{https://www.science.org/doi/abs/10.1126/sciadv.abe7953}.

\bibitem[Hoos et~al.(2021)Hoos, Hutter, and Leyton-Brown]{Hoos2021}
Hoos, H.~H., Hutter, F., and Leyton-Brown, K.
\newblock \emph{Chapter 12. Automated Configuration and Selection of SAT Solvers}.
\newblock IOS Press, February 2021.
\newblock \doi{10.3233/faia200995}.
\newblock URL \url{http://dx.doi.org/10.3233/FAIA200995}.

\bibitem[Hutter et~al.(2007)Hutter, Hoos, and Stützle]{Hutter2007ParamILS}
Hutter, F., Hoos, H., and Stützle, T.
\newblock Automatic algorithm configuration based on local search.
\newblock 01 2007.

\bibitem[Hutter et~al.(2009)Hutter, Hoos, Leyton-Brown, and St{\"u}tzle]{Hutter2009}
Hutter, F., Hoos, H.~H., Leyton-Brown, K., and St{\"u}tzle, T.
\newblock Paramils: an automatic algorithm configuration framework.
\newblock \emph{Journal of artificial intelligence research}, 36:\penalty0 267--306, 2009.

\bibitem[Hutter et~al.(2011)Hutter, Hoos, and Leyton-Brown]{Hutter2011SATTune}
Hutter, F., Hoos, H.~H., and Leyton-Brown, K.
\newblock Sequential model-based optimization for general algorithm configuration.
\newblock In Coello, C. A.~C. (ed.), \emph{Learning and Intelligent Optimization}, pp.\  507--523, Berlin, Heidelberg, 2011. Springer Berlin Heidelberg.
\newblock ISBN 978-3-642-25566-3.

\bibitem[Jamieson et~al.(2012)Jamieson, Nowak, and Recht]{Jamieson2012}
Jamieson, K.~G., Nowak, R., and Recht, B.
\newblock Query complexity of derivative-free optimization.
\newblock \emph{Advances in Neural Information Processing Systems}, 25, 2012.

\bibitem[Kalinin et~al.(2023)Kalinin, Mourgias-Alexandris, Ballani, Berloff, Clegg, Cletheroe, Gkantsidis, Haller, Lyutsarev, Parmigiani, Pickup, and Rowstron]{Kalinin2023}
Kalinin, K.~P., Mourgias-Alexandris, G., Ballani, H., Berloff, N.~G., Clegg, J.~H., Cletheroe, D., Gkantsidis, C., Haller, I., Lyutsarev, V., Parmigiani, F., Pickup, L., and Rowstron, A.
\newblock Analog iterative machine (aim): using light to solve quadratic optimization problems with mixed variables, 2023.

\bibitem[Karp(2010)]{Karp2010}
Karp, R.~M.
\newblock \emph{Reducibility among combinatorial problems}.
\newblock Springer, 2010.

\bibitem[Khudabukhsh et~al.(2009)Khudabukhsh, Xu, Hoos, and Leyton-Brown]{Khudabukhsh2009}
Khudabukhsh, A., Xu, L., Hoos, H., and Leyton-Brown, K.
\newblock Satenstein: Automatically building local search sat solvers from components.
\newblock volume 232, pp.\  517--524, 01 2009.
\newblock \doi{10.1016/j.artint.2015.11.002}.

\bibitem[KhudaBukhsh et~al.(2016)KhudaBukhsh, Xu, Hoos, and Leyton-Brown]{Khudabukhsh2016}
KhudaBukhsh, A.~R., Xu, L., Hoos, H.~H., and Leyton-Brown, K.
\newblock Satenstein: Automatically building local search sat solvers from components.
\newblock \emph{Artificial Intelligence}, 232:\penalty0 20--42, 2016.

\bibitem[Kim \& Zhang(2010)Kim and Zhang]{Kim2010}
Kim, S. and Zhang, D.
\newblock Convergence properties of direct search methods for stochastic optimization.
\newblock In \emph{Proceedings of the 2010 Winter Simulation Conference}, pp.\  1003--1011, 2010.
\newblock \doi{10.1109/WSC.2010.5679089}.

\bibitem[Kingma \& Ba(2017)Kingma and Ba]{ADAM}
Kingma, D.~P. and Ba, J.
\newblock Adam: A method for stochastic optimization, 2017.

\bibitem[Kolda et~al.(2003)Kolda, Lewis, and Torczon]{Kolda2003}
Kolda, T.~G., Lewis, R.~M., and Torczon, V.
\newblock Optimization by direct search: New perspectives on some classical and modern methods.
\newblock \emph{SIAM Review}, 45\penalty0 (3):\penalty0 385--482, 2003.
\newblock \doi{10.1137/S003614450242889}.
\newblock URL \url{https://doi.org/10.1137/S003614450242889}.

\bibitem[Kunstner et~al.(2023)Kunstner, Chen, Lavington, and Schmidt]{Kunstner2023}
Kunstner, F., Chen, J., Lavington, J.~W., and Schmidt, M.
\newblock Noise is not the main factor behind the gap between sgd and adam on transformers, but sign descent might be.
\newblock \emph{arXiv preprint arXiv:2304.13960}, 2023.

\bibitem[Lan \& Zhou(2018)Lan and Zhou]{lan2018optimal}
Lan, G. and Zhou, Y.
\newblock An optimal randomized incremental gradient method.
\newblock \emph{Mathematical programming}, 171:\penalty0 167--215, 2018.

\bibitem[Larson et~al.(2019)Larson, Menickelly, and Wild]{Larson2019}
Larson, J., Menickelly, M., and Wild, S.~M.
\newblock Derivative-free optimization methods.
\newblock \emph{Acta Numerica}, 28:\penalty0 287–404, 2019.
\newblock \doi{10.1017/S0962492919000060}.

\bibitem[Leleu et~al.(2019)Leleu, Yamamoto, McMahon, and Aihara]{Leleu2019}
Leleu, T., Yamamoto, Y., McMahon, P., and Aihara, K.
\newblock Destabilization of local minima in analog spin systems by correction of amplitude heterogeneity.
\newblock \emph{Physical Review Letters}, 122, 02 2019.
\newblock \doi{10.1103/PhysRevLett.122.040607}.

\bibitem[Leleu et~al.(2021)Leleu, Khoyratee, Levi, Hamerly, Kohno, and Aihara]{Leleu2021}
Leleu, T., Khoyratee, F., Levi, T., Hamerly, R., Kohno, T., and Aihara, K.
\newblock Scaling advantage of chaotic amplitude control for high-performance combinatorial optimization.
\newblock \emph{Communications Physics}, 4\penalty0 (1):\penalty0 266, 2021.
\newblock \doi{10.1038/s42005-021-00768-0}.
\newblock URL \url{https://doi.org/10.1038/s42005-021-00768-0}.

\bibitem[Lin \& Kernighan(1973)Lin and Kernighan]{LKHeuristic}
Lin, S. and Kernighan, B.~W.
\newblock An effective heuristic algorithm for the traveling-salesman problem.
\newblock \emph{Oper. Res.}, 21:\penalty0 498--516, 1973.
\newblock URL \url{https://api.semanticscholar.org/CorpusID:33245458}.

\bibitem[Liu et~al.(2023)Liu, Li, Hall, Liang, and Ma]{Liu2023sophia}
Liu, H., Li, Z., Hall, D., Liang, P., and Ma, T.
\newblock Sophia: A scalable stochastic second-order optimizer for language model pre-training.
\newblock \emph{arXiv preprint arXiv:2305.14342}, 2023.

\bibitem[Liu et~al.(2020)Liu, Liu, Gao, Chen, and Han]{Liu2020}
Liu, L., Liu, X., Gao, J., Chen, W., and Han, J.
\newblock Understanding the difficulty of training transformers.
\newblock \emph{arXiv preprint arXiv:2004.08249}, 2020.

\bibitem[Liu et~al.(2018)Liu, Kailkhura, Chen, Ting, Chang, and Amini]{Liu2018}
Liu, S., Kailkhura, B., Chen, P.-Y., Ting, P., Chang, S., and Amini, L.
\newblock Zeroth-order stochastic variance reduction for nonconvex optimization.
\newblock \emph{Advances in Neural Information Processing Systems}, 31, 2018.

\bibitem[Loshchilov \& Hutter(2017)Loshchilov and Hutter]{oshchilov2017decoupled}
Loshchilov, I. and Hutter, F.
\newblock Decoupled weight decay regularization.
\newblock \emph{arXiv preprint arXiv:1711.05101}, 2017.

\bibitem[Nelder \& Mead(1965)Nelder and Mead]{Nelder1965}
Nelder, J.~A. and Mead, R.
\newblock {A Simplex Method for Function Minimization}.
\newblock \emph{The Computer Journal}, 7\penalty0 (4):\penalty0 308--313, 01 1965.
\newblock ISSN 0010-4620.
\newblock \doi{10.1093/comjnl/7.4.308}.
\newblock URL \url{https://doi.org/10.1093/comjnl/7.4.308}.

\bibitem[Nesterov \& Spokoiny(2017)Nesterov and Spokoiny]{Nesterov2017}
Nesterov, Y. and Spokoiny, V.~G.
\newblock Random gradient-free minimization of convex functions.
\newblock \emph{Foundations of Computational Mathematics}, 17:\penalty0 527--566, 2017.
\newblock URL \url{https://api.semanticscholar.org/CorpusID:2147817}.

\bibitem[Ortega \& Rheinboldt(2000)Ortega and Rheinboldt]{Ortega2000}
Ortega, J.~M. and Rheinboldt, W.~C.
\newblock \emph{Iterative solution of nonlinear equations in several variables}.
\newblock SIAM, 2000.

\bibitem[Parisi(1980)]{Parisi1980}
Parisi, G.
\newblock The order parameter for spin glasses: a function on the interval 0-1.
\newblock \emph{Journal of Physics A: Mathematical and General}, 13\penalty0 (3):\penalty0 1101, mar 1980.
\newblock \doi{10.1088/0305-4470/13/3/042}.
\newblock URL \url{https://dx.doi.org/10.1088/0305-4470/13/3/042}.

\bibitem[Parizy et~al.(2023)Parizy, Kakuko, and Togawa]{Parizy2023}
Parizy, M., Kakuko, N., and Togawa, N.
\newblock Fast hyperparameter tuning for ising machines.
\newblock In \emph{2023 IEEE International Conference on Consumer Electronics (ICCE)}, pp.\  1--6, 2023.
\newblock \doi{10.1109/ICCE56470.2023.10043382}.

\bibitem[Pincus(1970)]{SA}
Pincus, M.
\newblock Letter to the editor—a monte carlo method for the approximate solution of certain types of constrained optimization problems.
\newblock \emph{Operations Research}, 18\penalty0 (6):\penalty0 1225--1228, 1970.
\newblock \doi{10.1287/opre.18.6.1225}.
\newblock URL \url{https://doi.org/10.1287/opre.18.6.1225}.

\bibitem[Reifenstein et~al.(2021)Reifenstein, Kako, Khoyratee, Leleu, and Yamamoto]{Reifenstein2021}
Reifenstein, S., Kako, S., Khoyratee, F., Leleu, T., and Yamamoto, Y.
\newblock Coherent ising machines with optical error correction circuits.
\newblock \emph{Advanced Quantum Technologies}, 4\penalty0 (11):\penalty0 2100077, 2021.
\newblock \doi{https://doi.org/10.1002/qute.202100077}.
\newblock URL \url{https://onlinelibrary.wiley.com/doi/abs/10.1002/qute.202100077}.

\bibitem[Reifenstein et~al.(2023)Reifenstein, Leleu, McKenna, Jankowski, Suh, Ng, Khoyratee, Toroczkai, and Yamamoto]{Reifenstein2023}
Reifenstein, S., Leleu, T., McKenna, T., Jankowski, M., Suh, M.-G., Ng, E., Khoyratee, F., Toroczkai, Z., and Yamamoto, Y.
\newblock Coherent sat solvers: a tutorial.
\newblock \emph{Adv. Opt. Photon.}, 15\penalty0 (2):\penalty0 385--441, Jun 2023.
\newblock \doi{10.1364/AOP.475823}.
\newblock URL \url{https://opg.optica.org/aop/abstract.cfm?URI=aop-15-2-385}.

\bibitem[Robbins \& Monro(1951)Robbins and Monro]{Robbins1951}
Robbins, H. and Monro, S.
\newblock {A Stochastic Approximation Method}.
\newblock \emph{The Annals of Mathematical Statistics}, 22\penalty0 (3):\penalty0 400 -- 407, 1951.
\newblock \doi{10.1214/aoms/1177729586}.
\newblock URL \url{https://doi.org/10.1214/aoms/1177729586}.

\bibitem[Sagun et~al.(2016)Sagun, Bottou, and LeCun]{Sagun2016}
Sagun, L., Bottou, L., and LeCun, Y.
\newblock Eigenvalues of the hessian in deep learning: Singularity and beyond.
\newblock \emph{arXiv preprint arXiv:1611.07476}, 2016.

\bibitem[Salimans et~al.(2017)Salimans, Ho, Chen, Sidor, and Sutskever]{Salimans2017}
Salimans, T., Ho, J., Chen, X., Sidor, S., and Sutskever, I.
\newblock Evolution strategies as a scalable alternative to reinforcement learning.
\newblock \emph{arXiv preprint arXiv:1703.03864}, 2017.

\bibitem[Schaul et~al.(2013)Schaul, Zhang, and LeCun]{Schaul2013}
Schaul, T., Zhang, S., and LeCun, Y.
\newblock No more pesky learning rates.
\newblock In \emph{International conference on machine learning}, pp.\  343--351. PMLR, 2013.

\bibitem[Schuetz et~al.(2022)Schuetz, Brubaker, and Katzgraber]{Schuetz2022}
Schuetz, M.~J., Brubaker, J.~K., and Katzgraber, H.~G.
\newblock Combinatorial optimization with physics-inspired graph neural networks.
\newblock \emph{Nature Machine Intelligence}, 4\penalty0 (4):\penalty0 367--377, 2022.

\bibitem[Schulman et~al.(2015)Schulman, Levine, Abbeel, Jordan, and Moritz]{Schulman2015}
Schulman, J., Levine, S., Abbeel, P., Jordan, M., and Moritz, P.
\newblock Trust region policy optimization.
\newblock In \emph{International conference on machine learning}, pp.\  1889--1897. PMLR, 2015.

\bibitem[Selman et~al.(1992)Selman, Levesque, and Mitchell]{GSAT}
Selman, B., Levesque, H., and Mitchell, D.
\newblock A new method for solving hard satisfiability problems.
\newblock In \emph{Proceedings of the Tenth National Conference on Artificial Intelligence}, AAAI'92, pp.\  440–446. AAAI Press, 1992.
\newblock ISBN 0262510634.

\bibitem[Selsam et~al.(2019)Selsam, Lamm, Bünz, Liang, de~Moura, and Dill]{Selsam2019}
Selsam, D., Lamm, M., Bünz, B., Liang, P., de~Moura, L., and Dill, D.~L.
\newblock Learning a sat solver from single-bit supervision, 2019.

\bibitem[Shamir(2015)]{Shamir2015}
Shamir, O.
\newblock An optimal algorithm for bandit and zero-order convex optimization with two-point feedback.
\newblock \emph{ArXiv}, abs/1507.08752, 2015.
\newblock URL \url{https://api.semanticscholar.org/CorpusID:2541603}.

\bibitem[Snoek et~al.(2012)Snoek, Larochelle, and Adams]{Snoek2012}
Snoek, J., Larochelle, H., and Adams, R.~P.
\newblock Practical bayesian optimization of machine learning algorithms.
\newblock \emph{Advances in neural information processing systems}, 25, 2012.

\bibitem[Spall(1998)]{Spall1998}
Spall, J.
\newblock Implementation of the simultaneous perturbation algorithm for stochastic optimization.
\newblock \emph{IEEE Transactions on Aerospace and Electronic Systems}, 34\penalty0 (3):\penalty0 817--823, 1998.
\newblock \doi{10.1109/7.705889}.

\bibitem[Sun \& Nocedal(2022)Sun and Nocedal]{Sun2022}
Sun, S. and Nocedal, J.
\newblock A trust region method for the optimization of noisy functions, 2022.

\bibitem[Taassob et~al.(2023)Taassob, Venturelli, and Lott]{Taassob2023}
Taassob, A., Venturelli, D., and Lott, P.~A.
\newblock Neural deep operator networks representation of coherent ising machine dynamics.
\newblock In \emph{Machine Learning with New Compute Paradigms}, 2023.

\bibitem[Wang et~al.(2013)Wang, Marandi, Wen, Byer, and Yamamoto]{Wang2013}
Wang, Z., Marandi, A., Wen, K., Byer, R.~L., and Yamamoto, Y.
\newblock Coherent ising machine based on degenerate optical parametric oscillators.
\newblock \emph{Phys. Rev. A}, 88:\penalty0 063853, Dec 2013.
\newblock \doi{10.1103/PhysRevA.88.063853}.
\newblock URL \url{https://link.aps.org/doi/10.1103/PhysRevA.88.063853}.

\bibitem[Yamamoto et~al.(2017)Yamamoto, Aihara, Leleu, Kawarabayashi, Kako, Fejer, Inoue, and Takesue]{Yamamoto2017}
Yamamoto, Y., Aihara, K., Leleu, T., Kawarabayashi, K.-i., Kako, S., Fejer, M., Inoue, K., and Takesue, H.
\newblock Coherent ising machines---optical neural networks operating at the quantum limit.
\newblock \emph{npj Quantum Information}, 3\penalty0 (1):\penalty0 49, 2017.
\newblock \doi{10.1038/s41534-017-0048-9}.
\newblock URL \url{https://doi.org/10.1038/s41534-017-0048-9}.

\bibitem[Yao et~al.(2020)Yao, Gholami, Keutzer, and Mahoney]{Yao2020}
Yao, Z., Gholami, A., Keutzer, K., and Mahoney, M.~W.
\newblock Pyhessian: Neural networks through the lens of the hessian.
\newblock In \emph{2020 IEEE international conference on big data (Big data)}, pp.\  581--590. IEEE, 2020.

\bibitem[Yao et~al.(2021)Yao, Gholami, Shen, Mustafa, Keutzer, and Mahoney]{Yao2021adahessian}
Yao, Z., Gholami, A., Shen, S., Mustafa, M., Keutzer, K., and Mahoney, M.
\newblock Adahessian: An adaptive second order optimizer for machine learning.
\newblock In \emph{proceedings of the AAAI conference on artificial intelligence}, volume~35, pp.\  10665--10673, 2021.

\end{thebibliography}
\bibliographystyle{icml2024}
\end{document}